\documentclass[journal,twoside,web]{ieeecolor}
\usepackage{generic}
\usepackage{amsmath,amsfonts}
\usepackage{algorithmic}
\usepackage{algorithm}
\usepackage{hyperref}
\hypersetup{hidelinks=true}
\usepackage{array}
\usepackage[caption=false,font=footnotesize,labelfont=sf,textfont=sf]{subfig}
\usepackage{textcomp}
\usepackage{stfloats}
\usepackage{url}
\usepackage{verbatim}
\usepackage{graphicx}
\usepackage{cite}
\usepackage{wrapfig}
\usepackage{comment}
\usepackage{tablefootnote}
\usepackage{threeparttable}
\usepackage{multirow}
\usepackage{tabularx}
\usepackage{booktabs}
\usepackage{pifont}
\newcommand{\cmark}{\textcolor{Green}{\ding{51}}}%
\newcommand{\xmark}{\textcolor{Red}{\ding{55}}}%
\newcommand{\qmark}{\textcolor{Orange}{\textbf{?}}}%
\usepackage[dvipsnames]{xcolor}
\usepackage{colortbl}
\definecolor{Gray}{gray}{0.85}
\let\labelindent\relax
\usepackage[inline]{enumitem}
\newenvironment{tableitemize}
{ \begin{minipage}[t]{\linewidth} \vspace{-10pt} \begin{itemize}[leftmargin=10pt] \vspace{5pt}}
{  \vspace{5pt} \end{itemize} \end{minipage}   }

\begin{document}

\title{Smart Pressure e-Mat for Human Sleeping Posture and Dynamic Activity Recognition}

\author{Liangqi Yuan,~\IEEEmembership{Student Member,~IEEE,}
        Yuan Wei, \IEEEmembership{Student Member,~IEEE}, 
        and Jia Li, \IEEEmembership{Senior Member,~IEEE}
\thanks{This research is supported by the AFOSR grant FA9550-21-1-0224.}
\thanks{Liangqi Yuan and Yuan Wei are with the School of Electrical and Computer Engineering, Purdue University, West Lafayette, IN 47907, USA (e-mail: liangqiy@purdue.edu; wei431@purdue.edu).}
\thanks{Jia Li is with the Department of Electrical and Computer Engineering, Oakland University, Rochester, MI 48309, USA (e-mail: li4@oakland.edu).}}

\def\BibTeX{{\rm B\kern-.05em{\sc i\kern-.025em b}\kern-.08em
    T\kern-.1667em\lower.7ex\hbox{E}\kern-.125emX}}
\markboth{\hskip25pc IEEE TRANSACTIONS AND JOURNALS TEMPLATE}
{Author \MakeLowercase{\textit{et al.}}: Title}

\maketitle

\begin{abstract}
With the emphasis on healthcare, early childhood education, and fitness, non-invasive measurement and recognition methods have received more attention. Pressure sensing has been extensively studied because of its advantages of simple structure, easy access, visualization application, and harmlessness. This paper introduces a Smart Pressure e-Mat (SPeM) system based on piezoresistive material, Velostat, for human monitoring applications, including recognition of sleeping postures, sports, and yoga. After a subsystem scans the e-mat readings and processes the signal, it generates a pressure image stream. Deep neural networks (DNNs) are used to fit and train the pressure image stream and recognize the corresponding human behavior. Four sleeping postures and 13 dynamic activities inspired by Nintendo Switch Ring Fit Adventure (RFA) are used as a preliminary validation of the proposed SPeM system. The SPeM system achieves high accuracies in both applications, demonstrating the high accuracy and generalizability of the models. Compared with other pressure sensor-based systems, SPeM possesses more flexible applications and commercial application prospects, with reliable, robust, and repeatable properties.
\end{abstract}


\section{Introduction}
\label{sec:introduction}

\IEEEPARstart{H}{UMAN} activity recognition (HAR) aims to identify activities through a snapshot of observations of the subject's behavior and environmental conditions. HAR research has been used most extensively in healthcare \cite{5,12}, sports \cite{32}, human-computer interaction (HCI) \cite{2}, security \cite{14}, and robotics \cite{3}. The mainstream mechanism of HAR is divided into wearable sensors and external devices. Wearable sensors are widely used in the HAR field \cite{yuan2024fedmfs,yuan2024communication,attal2015physical,nweke2018deep,wang2019survey,luo2021learning}, due to their fit to the human body, valid signal, compact sensor size, better spatial freedom, and the ability to work in various complex environments. However, wearable sensors suffer from the drawback of requiring the subject to wear or mount multiple sensors on different parts of the body. External devices are deployed to observe, detect, recognize, and segment human features in the scenario. Vision-based recognition methods involve computer vision (CV) \cite{21,yuan2023federated,yuan2023peer}, usually using RGB, depth, infrared, and thermal imaging cameras. The basic idea is to take the whole image as input and show the pixel coordinates of key points of the body. Due to non-contact measurement, the wider corresponding spectral range, and the ability to work stably for a long time for vision inspection, CV systems are widely used in industry, agriculture, defense, transportation, medicine, entertainment, and so on \cite{zheng2018multispectral}. Vision-based HAR has some inherent limits, for example, data acquisition is seriously affected by lighting or temperature of the environment \cite{13} and involves privacy concerns \cite{15}. Other types of external HAR methods, such as radio frequency (RF) sensors \cite{yuan2022interpretable,18,mu2023human}, can accurately classify human subjects and activities, but most of them are hard to visualize human behavior. 

\begin{figure}[t]
\centerline{\includegraphics[width=\columnwidth]{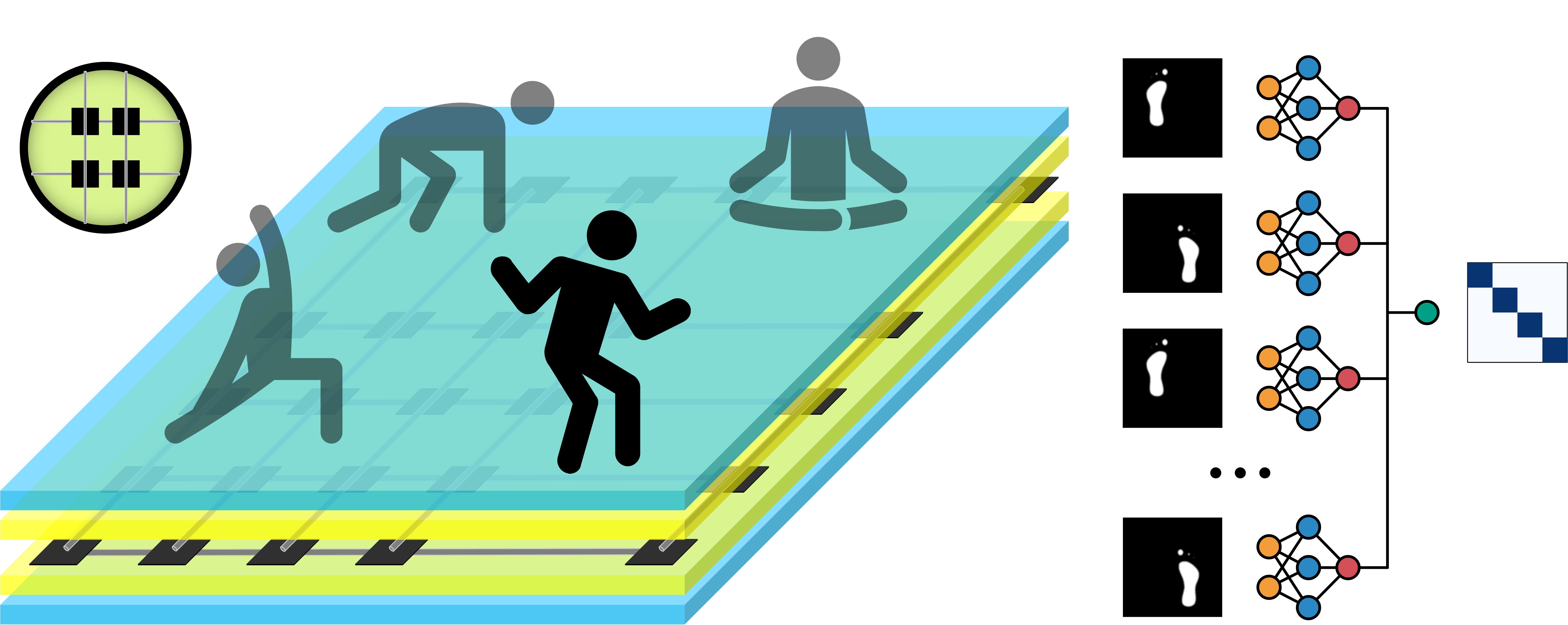}}
\caption{\textbf{Schematic Illustration of the SPeM System for Human Activity Recognition.} Different human activities generate distinct pressure distributions, which in turn result in varied resistance distributions. These resistance distributions subsequently give rise to diverse voltage distributions, enabling the generation of a stream of pressure images classified into different activities by a deep neural network.}
\label{Fig. SPeM}
\end{figure}

\subsection{Velostat-Based Application}
\label{Sec. Velostat-Based Application}

Inspired by the mechanism that humans have sensory functions when they are in direct contact with the external environment \cite{4}, tactile sensing is widely used in the fields of intelligent HCI technology and biomedical monitoring, which is an essential means of data acquisition, analysis, and control of machines to perceive the external environment. Pressure sensing is the tactile sensing mode used in this paper. It has diversified production methods, strong versatility, and convenient configuration, ensuring lower production costs and more uses. Pressure sensors have multiple subcategories, including capacitive, piezoelectric, optical, piezoresistive, etc. \cite{chi2018recent}. Velostat \cite{23}, also known as Linqstat, is a packaging material made of polymer foil impregnated with carbon black to make it conductive. Velostat-based piezoresistive pressure sensor arrays have been widely studied and used in recent years. Despite its non-ideal electrical properties and the crosstalk observed in numerous recent studies \cite{yuan2021velostat,medrano2019circuit,dzedzickis2020polyethylene}, which can reduce its responsiveness, some studies \cite{cao2024crosstalk} have found that adding diodes to the circuit can mitigate the effects of crosstalk. However, this approach may lead to increased fragility and discomfort. Nevertheless, despite its electrical limitations, its low price, flexibility, and scalability have garnered attention, especially in human sensing applications \cite{cao2024mortise}. In this paper, we select a Velostat sensor array as a tool to collect pressure distribution and complete HAR with deep learning (DL).

\begin{table*}[t]
\renewcommand*{\arraystretch}{1.5}
\begin{threeparttable}
\caption{Comparison with the State-of-the-art Pressure Sensor Array Technology}
\label{Table Comparison}
\centering
\begin{tabularx}{\linewidth}{lXlp{60pt}p{40pt}ll}
\toprule
\textbf{Literature} & \textbf{Tasks} & \textbf{Material} & \textbf{Length $\times$ Width (m $\times$ m)} & \textbf{Resolution ($\text{mm}^2/\text{pixel}$)} & \textbf{Price (USD)} & \textbf{Foldable} \\
\specialrule{\lightrulewidth}{0.2 em}{0pt}
\rowcolor{Gray}
Tactilus \cite{Tactilus} & N/A (Commercial Sensor System) & \qmark & 1.830 $\times$ 0.603 & 31.0 $\times$ 37.7 & 7200 & \qmark \\

Tekscan 5400N \cite{Tekscan} & N/A (Commercial Sensor System) & \qmark & 1.060 $\times$ 0.640 & 17.0 $\times$ 17.0 & \qmark & \qmark \\
\rowcolor{Gray}
Fatema \textit{et al.} \cite{fatema2021low} & \begin{tableitemize} \item Object Movement Detection \end{tableitemize} & Velostat & 0.300 $\times$ 0.300 & 75 $\times$ 75 & \qmark & \xmark \\

Yuan \textit{et al.} \cite{yuan2021velostat} & \begin{tableitemize} \item Object Recognition \end{tableitemize} & Velostat & 0.140 $\times$ 0.140 & 5 $\times$ 5 & 25 & \xmark \\
\rowcolor{Gray}
Sun \textit{et al.} \cite{sun2017compressed} & \begin{tableitemize} \item Respiratory Waveform Reconstruction \end{tableitemize} & Velostat & 0.385 $\times$ 0.360 & 9 $\times$ 9 & \qmark & \cmark \\

Wan \textit{et al.} \cite{wan2021hip} & \begin{tableitemize} \item Sitting Posture Recognition \end{tableitemize} & \qmark & 0.365 $\times$ 0.365 & 11.5 $\times$ 11.5 & 150 & \xmark \\
\rowcolor{Gray}
Yuan \textit{et al.} \cite{yuan2021smart} & \begin{tableitemize} \item Sitting Posture Recognition \end{tableitemize} & Velostat & 0.381 $\times$ 0.381 & 11.3 $\times$ 11.3 & 45 & \cmark \\
Cao \textit{et al.} \cite{cao2024crosstalk} & \begin{tableitemize} \item Sitting Posture Recognition \end{tableitemize} & Velostat & 0.32 $\times$ 0.22 & 22.9 $\times$ 24.4 & \qmark & \xmark \\

\rowcolor{Gray}
Tang \textit{et al.} \cite{34} & \begin{tableitemize} \item Sleeping Posture Recognition \end{tableitemize} & Velostat & 2.000 $\times$ 1.000 & 100 $\times$ 100 & 51 (sensor) & \xmark \\
Hu \textit{et al.} \cite{26} & \begin{tableitemize} \item Sleeping Posture Recognition \end{tableitemize} & Velostat & 1.800 $\times$ 0.900 & 54.5 $\times$ 27.2 & 130 (sensor) & \cmark \\

\rowcolor{Gray}
Hudec \textit{et al.} \cite{hudec2020smart} & \begin{tableitemize} \item Sleeping Posture Recognition \end{tableitemize} & Velostat & 0.900 $\times$ 0.900 & 100 $\times$ 100 & \qmark & \cmark \\
Sundholm \textit{et al.} \cite{sundholm2014smart} & \begin{tableitemize} \item Activity Pattern Recognition \end{tableitemize} & CarboTex & 0.800 $\times$ 0.800 & 10 $\times$ 10 & \qmark & \cmark \\

\rowcolor{Gray}
Zhou \textit{et al.} \cite{zhou2017carpet} & \begin{tableitemize} \item Subject Gait Identification \end{tableitemize} & CarboTex & 1.800 $\times$ 0.800 & 15 $\times$ 15 & \qmark & \cmark \\
Wicaksono \textit{et al.} \cite{wicaksono20223dknits} & \begin{tableitemize} \item Activity Pattern Recognition \end{tableitemize} & Polypyrrole & 0.450 $\times$ 0.450 & 25 $\times$ 25 & \qmark & \cmark \\
\rowcolor{Gray}
Cao \textit{et al.} \cite{cao2024mortise} & \begin{tableitemize} \item Activity Monitoring \end{tableitemize} & Velostat & 1.820 $\times$ 0.360 & 65 $\times$ 60 & \qmark & \xmark \\

\textbf{This Paper} & \begin{tableitemize} \item \textbf{Sleeping Posture Recognition} \item \textbf{Dynamic Activity Recognition} \end{tableitemize} & \textbf{Velostat} & \textbf{2.030 $\times$ 1.525} & \textbf{67.3 $\times$ 52.3} & \textbf{220} & \cmark \\
\bottomrule
\end{tabularx}
\begin{tablenotes}
\item \hspace{-8pt} \cmark {} Yes, \qmark {} No report, \xmark {} No.
\end{tablenotes}
\end{threeparttable}
\end{table*}

The pressure sensor array aims to convert the vertically oriented pressure input into a standard grayscale image. With DL algorithms such as convolutional neural networks (CNNs) that have excellent performance in image recognition and processing, the DL and pressure sensor array are able to classify contact objects with high accuracy. Tang \emph{et al.} \cite{34} designed a pressure mattress with Velostat material for the hospice care of the elderly. The internet of things (IoT) based solution used sensors to record patient posture-related data and transmitted it to the cloud for further processing. Hu \emph{et al.} \cite{26} developed an on-the-fly human sleep recognition system using a pressure-sensitive conductive sheet and a four-layer CNN architecture for sleep classification, with transfer learning to prevent overfitting and improve classification accuracy. Yuan \emph{et al.} \cite{yuan2021velostat} established an extensive object recognition to classify ten objects and performed a systematic material analysis and study of Velostat, including resistance sensitivity, quasi-static response, and crosstalk issues. Zhang \emph{et al.} \cite{33} focused on gait recognition using a combination of pressure signals and acceleration signals to compensate for the lack of data provided by a single sensor and transmitted the data to a computer for signal processing and building a \emph{k} nearest neighbor (\emph{k}NN) model to test the gait pattern recognition effect. Chen \emph{et al.} \cite{27} explored DL algorithms including ResNet50, InceptionV3, and MobileNet to identify differences in the response of walking speed to plantar pressure. Jun \emph{et al.} \cite{29} performed pathological gait classification, feeding the sequential skeleton and average foot pressure data into a recurrent neural network (RNN)-based encoding layers and CNN-based encoding layers, respectively. The method effectively extracted features, then the output features were connected and fed to a fully connected layer for classification. Ghzizal \emph{et al.} \cite{28} used transfer learning of a pre-trained CNN to classify patients with Parkinson's disease. Tactile perception is an important research direction in the field of robotics and artificial skin. We summarize the details of the state-of-the-art pressure sensor array technologies in Table \ref{Table Comparison}.

\subsection{Dynamic Pattern Recognition}
\label{Sec. Dynamic Pattern Recognition}

Pressure pattern in the physical world, especially tactile sensing in relation to humans, is a dynamic modality. Different people have different physical characteristics and behavioral habits, which affects the generation of pressure distribution. Meanwhile, the non-ideal properties of most pressure sensors, especially piezoresistive sensors, also cause non-linear effects on resistance and conductance. Traditional classifiers are difficult to handle such tasks. Therefore, it is a trend to use DL algorithms to classify a series of dynamic pressure images, known as a stream of pressure images. There are traceable solutions for image stream or video recognition. In the field of image sequence recognition, the convolutional recurrent neural network (CRNN) is used for the extraction of language text from the image \cite{shi2016end} and video classification \cite{yue2015beyond}, in which long short-term memory (LSTM) is used to integrate CNNs. It is also common practice for researchers to recognize dynamic pressure image sequences using integrated CNNs. Song \emph{et al.} \cite{song2022contact} utilized a similar CRNN architecture to identify four modes of flexible tactile sensors, including stroking, patting, kneading, and scratching. Sundaram \emph{et al.} \cite{7} demonstrated tactile gloves for object-grasping robotics, in which a convolutional layer was used to integrate CNNs and was used to recognize the type of objects and judge the response gestures of robotics. Therefore, the deep neural network (DNN) algorithm is used as the pressure pattern recognition method in this paper, and the performances of the three methods are discussed and compared.

\subsection{Motivation and Contribution}
\label{Sec. Motivation and Contribution}

Our motivation received inspiration and impetus from healthcare, sport, gaming, early childhood education, entertainment, etc. This work aims to introduce a flexible, non-invasive, portable, and inexpensive pressure sensor system into the above areas of application. On the one hand, pressure can be a Boolean feedback as a single-dimensional signal detection method. However, the pressure distribution formed by the human body on a pressure array can generate a visualized pressure image. Gumus \emph{et al.} \cite{gumus2022textile} presented a textile-based pressure sensor array that was capable of displaying the shape of objects, recognizing gestures, and early childhood education. Children could input and display numbers, arithmetic symbols, and English letters on an Android mobile phone by pressing them on the education platform. Wicaksono \textit{et al.} \cite{wicaksono20223dknits} developed a knitted intelligent textile mat to control the standing, walking, and running behavior of the characters in the Minecraft video game. 
Moreover, pressure sensor arrays have shown substantial potential in healthcare applications such as the prevention of bedsores and detection of falls, among others \cite{wai2009sleeping,aoki2014association,matar2019artificial}. This paper explores a wider range of applications, utilizing an electronic mat composed of a pressure sensor array to enable recognition of sleeping postures and various activities, such as running, exercising, yoga, among others. Through these explorations, our aim is to uncover its potential in healthcare applications and beyond.

In this paper, a pressure sensor-based system, Smart Pressure e-Mat (SPeM), is proposed for human sleeping posture and recognition of dynamic activity, as shown in Fig. \ref{Fig. SPeM}. SPeM consists of a pressure sensor array e-mat based on Velostat piezoresistive material and a DNN classifier. After demonstrating the fabrication of the pressure sensor e-mat, a DNN algorithm with three alternate architectures is designed to process the pressure image stream. SPeM collects two datasets, including four sleeping postures and 13 dynamic activities, which are used to train different DNN models, respectively. After evaluation and discussion of the experiment results, our SPeM system is considered a highly accurate, low-cost, and convenient human monitoring application. The contributions of this paper are as follows.
\begin{enumerate}
\item \textbf{\textit{Sensor Array Design:}} A Velostat sensor array-based SPeM is designed and established to generate human pressure patterns. SPeM is low in price, foldable, portable, flexible, and low weight, with a length of 2.030 meters, a width of 1.525 meters, a height of fewer than 0.001 meters, total weight is 2749 grams, its total price is 220 USD, and the resolution is $67.3 \times 52.3 \, \text{mm}^2/\text{pixel}$. We employ a specialized design method utilizing separated Velostat sensor elements, effectively reducing the consumption of Velostat material by 64\%.
\item \textbf{\textit{Sleeping Postures and Dynamic Activities Recognition:}} SPeM is employed in our experimental setup to monitor human sleeping postures and dynamic activities within domestic scenarios. We have gathered human pressure pattern datasets across two distinct environments, including four sleeping postures and 13 dynamic activities. These datasets comprise a total of 30,000 image stream samples, with each sample containing ten image frames.
\item \textbf{\textit{Pressure Image Stream Classification:}} DNN algorithms are utilized to recognize pressure images of sleeping postures and dynamic activities, achieving notable accuracy. Experimental results demonstrate that the proposed SPeM effectively captures the pressure modality generated by human postures/activities, offering a high-precision, comprehensive, and visualizable paradigm and instance.
\end{enumerate} 

The presentation of the work is as follows. In Section II, the methodology of this work is described. The experimental setups and results are demonstrated in Section III. Section IV discusses the deployment of the proposed system in the real world, while Section V concludes the paper and outlines future work.

\section{Methodologies}
\label{Sec. Methodologies}
\begin{figure}[h]
\centerline{\includegraphics[width=\columnwidth]{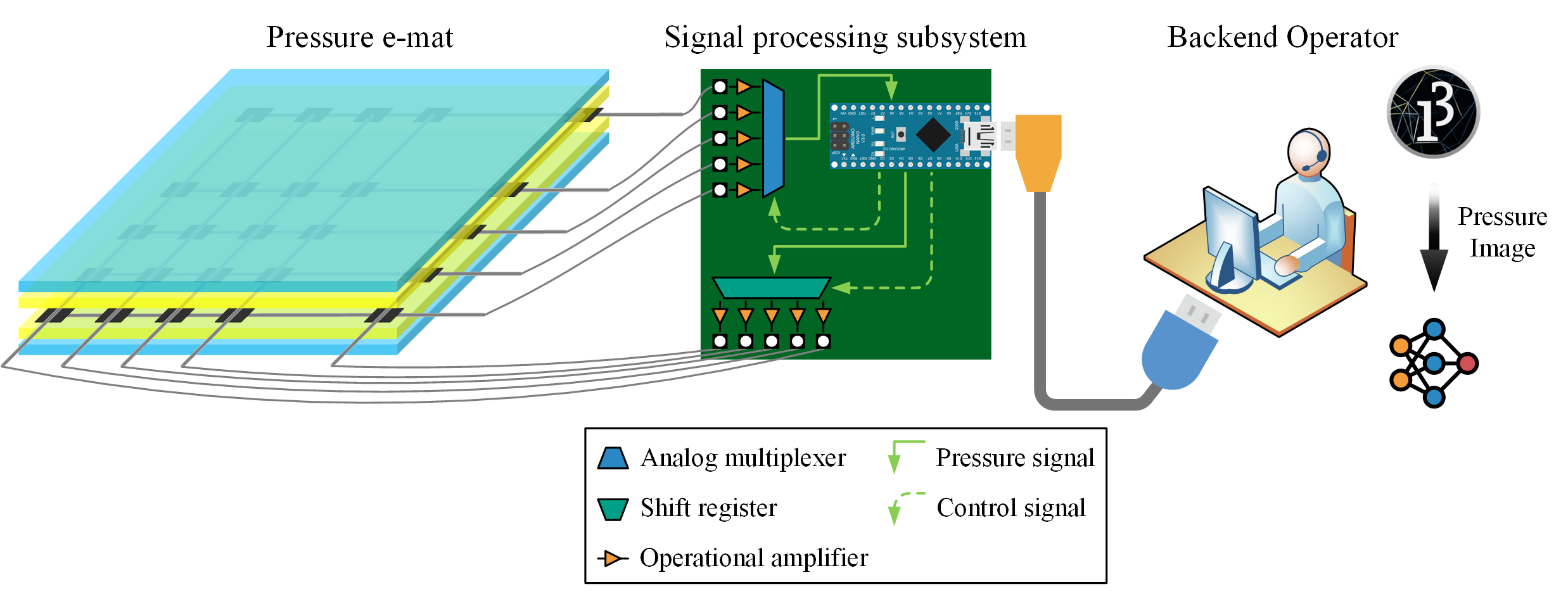}}
\caption{\textbf{Design Diagram of Proposed SPeM System.} The human pressure modality is represented as pressure distributions on the pressure e-mat. The signal processing subsystem generates a voltage distribution by scanning the interface of the pressure e-mat. The backend operator generates pressure images and utilizes a classifier for classification.}
\label{Fig. system}
\end{figure}

This section presents the proposed SPeM system, including the fabrication of the e-mat based on Velostat pressure sensor array and the design of the DNN algorithm. The schematic diagram of the proposed SPeM system design is presented in Fig. \ref{Fig. system}.

\subsection{Smart Pressure e-Mat System Design}
\label{Sec. Smart Pressure e-Mat System Design}

The proposed SPeM system comprises a pressure e-mat for sensing, a signal processing subsystem for calibration, scanning, and sampling, and a back-end for visualization and classification. The signal processing subsystem consists of a printed circuit board (PCB) integrated with analog multiplexers, shift registers, and operational amplifiers, and embedded with an Arduino Nano as the central signal processing unit. To generate pressure images in real-time for visualization and classification, a Processing program is run by the back-end operator, which utilizes the voltage distribution produced by the signal processing subsystem. Our previous work \cite{yuan2021velostat} served as a systematic feasibility study that validated the signal processing subsystem; refer to \cite{yuan2021velostat} for more detailed schematic circuit diagrams and other details. The human body exerts varying pressure on the pressure e-mat depending on different tasks, resulting in distinct resistance distributions. Arduino controls analog multiplexers and shift registers to scan the pressure mat line by line and obtain the corresponding analog voltage distribution. Therefore, the quality of the pressure image is based on the ability of the pressure cushion to completely, accurately, and reliably receive the human body pressure distribution.

\subsection{Fabrication of SPeM}
\label{Sec. Fabrication of SPeM}

\begin{figure}[t]
\centerline{\includegraphics[width=\columnwidth]{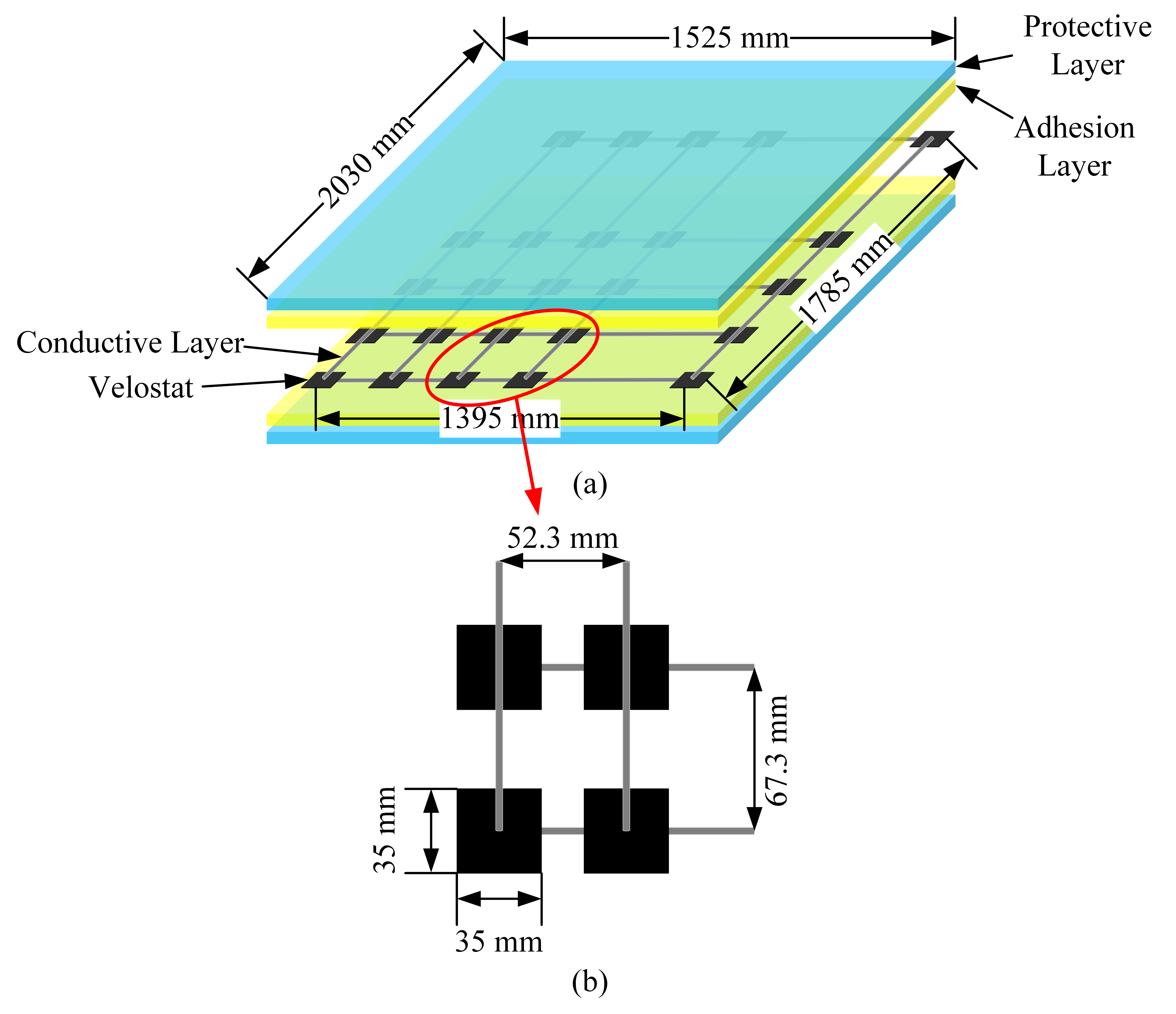}}
\caption{\textbf{Schematic Diagram of e-Mat.} (a) Structure, size, and sensing area. (b) Size and spacing of sensor elements.}
\label{Fig. Smart Pressure Mat}
\end{figure}

For the Velostat pressure sensor array, the pressure distribution represented by the pressure image is only related to the Velostat resistance at the intersections of the row and column of conductive lines, i.e., the pressure sensor elements, and the pressure distribution between elements is not sampled. Separated sensor elements, i.e., using cut Velostat instead of a whole piece of Velostat, is a viable approach to avoid waste of material in inactive areas, as shown in Fig. \ref{Fig. Smart Pressure Mat}. Separated sensor elements have three advantages: save material, avoid stray current flowing through the Velostat resistance material to adjacent elements, and avoid the deformation caused by human body pressure to affect adjacent elements through the material. The design approach of the separated Velostat sensor elements can reduce the use of Velostat material by 64$\%$. Meanwhile, separated sensor elements need to be distinguished from independent sensor elements. The independent sensor element is that each sensor element does not share the input and output circuits with other elements, while each row and each column of separated sensor elements share the input and output circuits, respectively. Therefore, compared with independent sensor elements, separated sensor elements have a minimal number of input and output ports, resulting in a simpler signal processing subsystem, but there will inevitably be crosstalk. Independent sensor elements are more suitable for the requirement of a small number of elements, while separated sensor elements are required in large sensor arrays.

\begin{figure}[t]
\centering
\subfloat[\label{Fig. Actual Mat}]{%
  \includegraphics[width=0.5\linewidth]{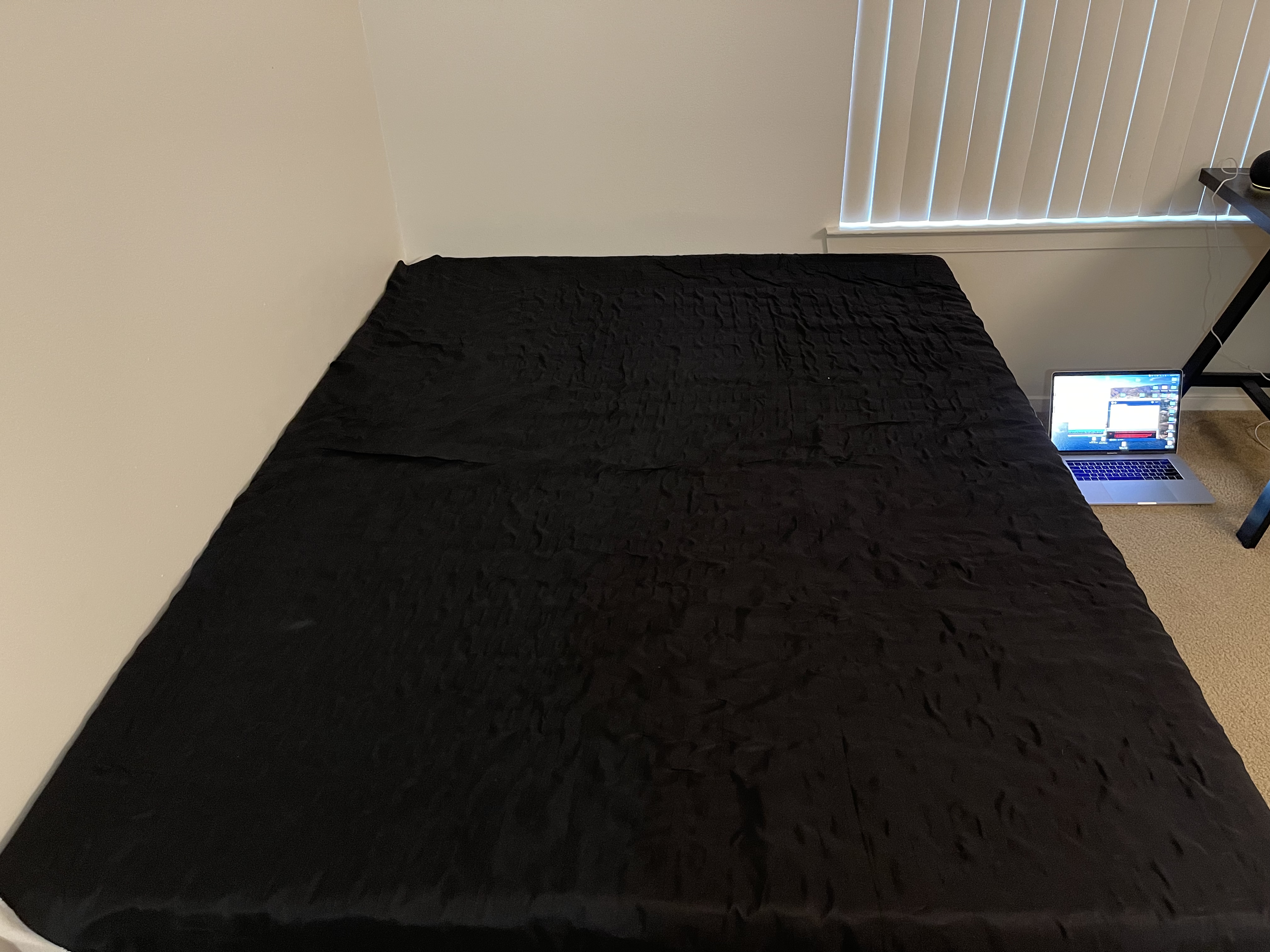}}
\hfill
\subfloat[\label{Fig. Folded Mat}]{%
  \includegraphics[width=0.5\linewidth]{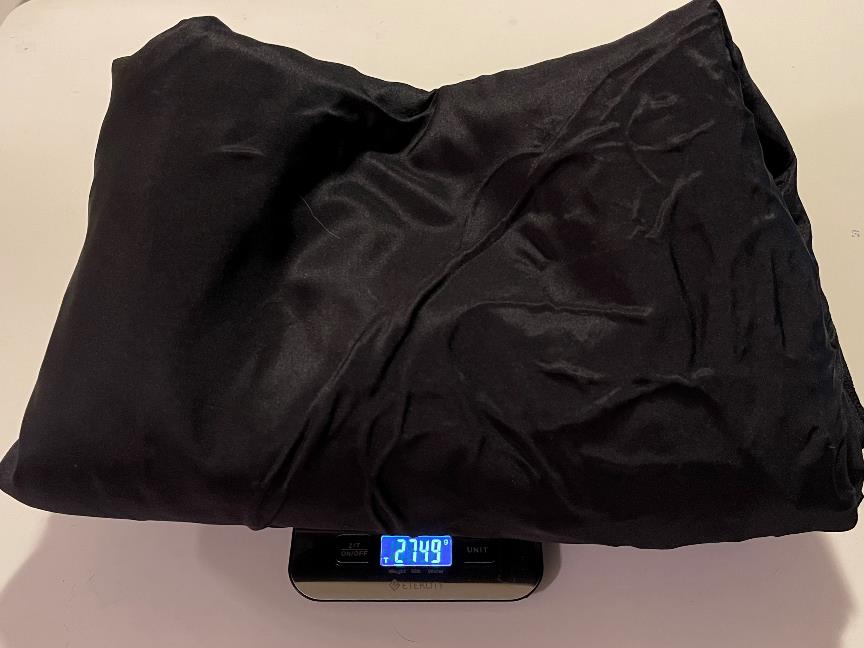}}
\caption{\textbf{Actual Image of e-Mat.} (a) The e-mat is placed on a Queen size mattress. (b) The e-mat is folded and placed on a digital weight scale, and the total weight is 2749 grams.}
\label{Fig. Actual Smart Pressure Mat} 
\end{figure}

Fig. \ref{Fig. Smart Pressure Mat}(a) shows the general structure and size of the e-mat, consisting of five layers, in which two adhesive layers are used to fix the upper and lower protective layers with the middle conductive layer, respectively. The protective layers are polyester due to their cheap, wrinkle-resistant, durable, and soft properties. The adhesive layers are acrylic, which is not only soft, but also firm and efficient for bonding multiple objects together, such as protective layers, conductive threads, and Velostat. The conductive threads are stainless steel fiber, which is famous for its role in conductive fabrics, and we have taken a fancy to its softness. Therefore, the soft property of the selected material is the core consideration, leading to the suitability of the fabricated e-mat for human-related applications. Unlike other Velostat-based pressure sensor arrays, the distances between sensor elements in the vertical and horizontal directions of the e-mat are not equidistant, as shown in Fig. \ref{Fig. Smart Pressure Mat}(b). In order to conform to the length and width of the human body when lying on the bed, the size of e-mat is set to be equal to the Queen size, and the actual sensing area is slightly smaller than the bed sheet. Fig. \ref{Fig. Actual Mat} shows the Smart Bed Sheet placed on a Queen-size mattress to collect data on subject sleeping posture. The fabricated Smart Bed Sheet is flexible, soft, and portable, and Fig. \ref{Fig. Folded Mat} shows its foldability and low weight. We summarize the parameters of the proposed SPeM in Table \ref{Table Parameters of SPeM}.

\begin{table}[h]
\caption{Parameters of SPeM}
\label{Table Parameters of SPeM}
\centering
\begin{tabular}{@{}lc@{}}
\toprule
\textbf{Item} & \textbf{Description} \\
\midrule
Length $\times$ Width (m $\times$ m) & 2.030 $\times$ 1.525 \\

Resolution ($\text{mm}^2/\text{pixel}$) & 67.3 $\times$ 52.3 \\

Weight (gram) & 2749 \\

Price (USD) & 220 \\

Image sample rate (Hz) & 2 \\

Image stream sample rate (Hz) & 0.2 \\

Image stream resolution (pixel) & 27 $\times$ 27 $\times$ 1 $\times$ 10 \\
\bottomrule
\end{tabular}
\end{table}

\subsection{Deep Neural Network for Image Stream}
\label{Sec. Neural Network for Image Stream}
In this paper, we advocate the use of an end-to-end DNN as a pressure image stream classifier to achieve both human sleeping posture and dynamic activity recognition. CNN is one of the leading solutions for image classification problems. However, how to handle the temporal relationship between images after 2D convolution is still an open problem. In pressure image classification tasks, the sample is 3D tensors $\mathcal{I} \in \mathbb{R}^{n \times m \times d}$ and the label is $\mathcal{Y} \in \mathbb{N}$, where $n$ is the image length, $m$ is the width, and $d$ is the channel. $\mathbb{R}$ and $\mathbb{N}$ are real numbers and natural numbers, respectively. The length $n$, width $m$, and channel $d$ of the pressure image are determined by the actual manufacture of the e-mat. Although the classification tasks proposed in this paper are based on an image stream, also known as video, the sample is a 4D tensor $\mathcal{V} \in \mathbb{R}^{n \times m \times d \times j}$, where $j$ is a customized number of images in the stream, depending on the application requirements. In this paper, the dimensions of the sample are $27 \times 27 \times 1 \times 10$. Considering the non-linear resistance characteristics of the Velostat material, the image stream generated by SPeM depends not only on the pressure distribution of human activity but also on the electrical sensitivity of the Velostat based on the resulting pressure distribution. Specifically, resistance sensitivity is higher when the activity generates greater pressure and the pressure is applied for a shorter time, and vice versa. Considering the recognition of human sleep posture with a large pressure distribution area and a lengthy application time, even if it is a static human activity, the stream still presents different pressure images due to the nonlinear resistance characteristics of Velostat. In addition to the above two factors of activity and Velostat nonlinear characteristics, possible secondary factors include the subject's personal habits, noise, etc.

\begin{figure*}[t]
\centerline{\includegraphics[width=0.8\linewidth]{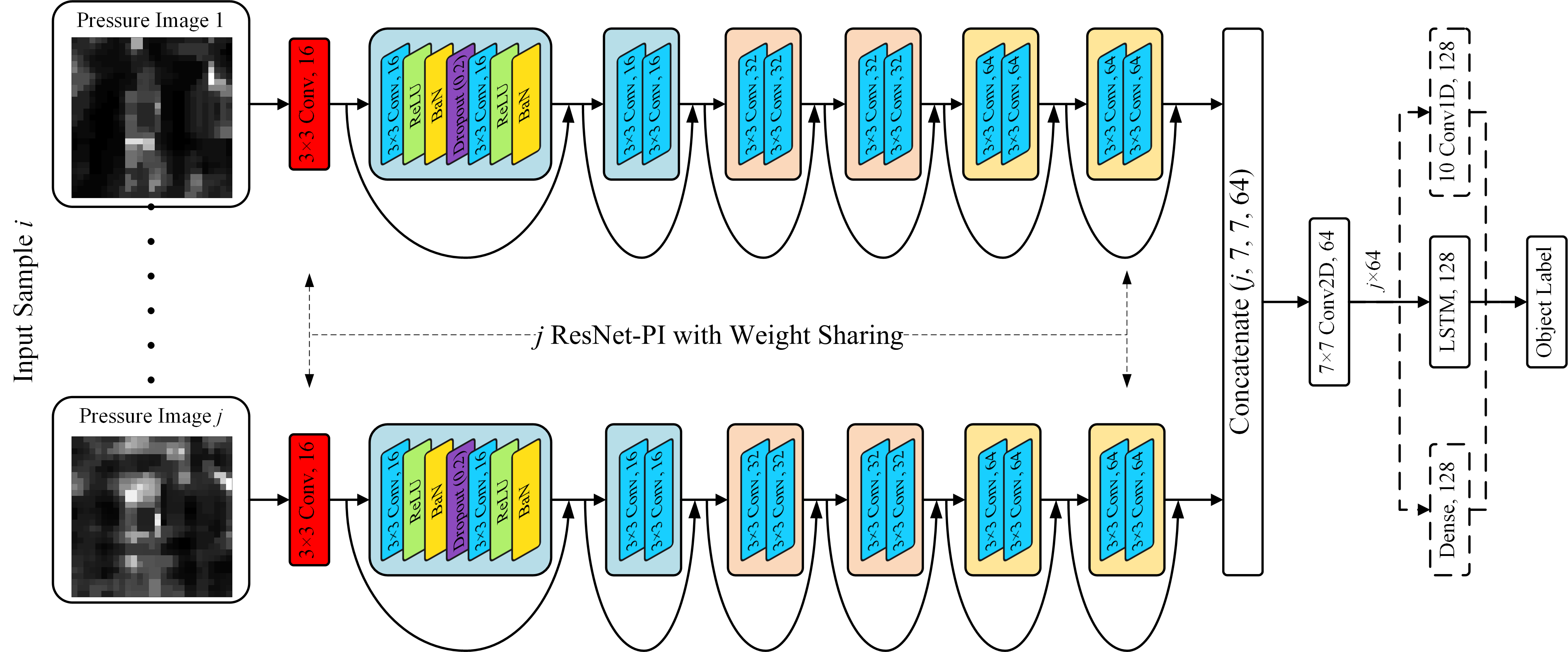}}
\caption{\textbf{Fusion of \textsc{ResNet-PI}s for Images with Temporal Relationship.} The DNN with three alternate architecture can be considered as 3D CNN, CRNN, and CDNN when concatenated with Conv, LSTM, or Dense, respectively.}
\label{Fig. CRNN}
\end{figure*}

Therefore, the critical thing is to find the temporal relationship of $j$ pressure images. Due to the excellent performance of CNN in image processing, it is used as a feature extractor to extract a high-level representation of human pressure patterns. Subsequently, the temporal relationships between images in a stream also need to be processed by neurons. Consequently, after $j$ CNNs are used to extract $j$ image tensors $\mathcal{I}$, a final layer processes the relationship between images in the stream, as a DNN classifier of the proposed SPeM system. Specifically, we consider \cite{yuan2021velostat} proposed CNN \textsc{ResNet-PI} as the pressure image feature extractor of DNN. \textsc{ResNet-PI} reduces the number of model parameters by removing one residual block from ResNet-18 \cite{he2016deep}. \textsc{ResNet-PI}, as a lightweight CNN, not only reduces computational complexity but also avoids overfitting pressure image data with a small number of features. Therefore, the DNN algorithm uses a temporal feature-level concatenation of \textsc{ResNet-PI}s, using pending convolution (Conv), long short-term memory (LSTM), or fully-connected (Dense) to process the temporal relationships. Fig. \ref{Fig. CRNN} shows the proposed DNN with three alternate architecture integrating \textsc{ResNet-PI}s.

The proposed DNN can adjust the number of sub-CNN according to the number $j$ of pressure images. Considering the need for recognition speed for human monitoring applications, the experiment in this paper considers a sample including $j=10$ pressure images, which results in a sampling time of five seconds for each sample. Considering that sub-CNNs are only used as feature extractors here, they can increase the learning efficiency and reduce the computational complexity by sharing weights. A Dropout layer is inserted into each ResNet block to further enhance the generalizability of the proposed DNN. The entire process of generating the pressure image stream dataset and training the DNN model by the SPeM system is elaborated in Algorithm \ref{Alg. SPeM System}.

\begin{algorithm}[h]
\small
\caption{Smart Pressure e-Mat (SPeM) System}
\label{Alg. SPeM System}
    {\bfseries Input:} Initial DNN model $\theta^0$, training epochs $E$, learning rate $\eta$, loss function $f$, Adam optimizer. \\
    {\bfseries Output:} Trained DNN model $\theta^\ast$

    \begin{algorithmic}[1]
    \item[] {\bfseries \# Scanning e-Mat to generate pressure image stream}
        \WHILE{data collection}
            \FOR{stream $ = 1, \dots, j$}
                \FOR{row $ = 1, \dots, n$}
                    \FOR{col $ = 1, \dots, m$}
                        \STATE Pressure image $\mathcal{I}^{\text{stream}}_{\text{row},\text{col}} $ $\gets$ 8-bit analog output
                    \ENDFOR
                \ENDFOR
                \STATE Generate pressure image $\mathcal{I}^{\text{stream}}$
                \STATE Pause 0.5 seconds
            \ENDFOR
            \STATE Concatenate $\boldsymbol{\mathcal{I}}$ and generate pressure image stream $\mathcal{V}$
            \STATE Append dataset $\xi$ with stream and label pair $\{\mathcal{V},\mathcal{Y}\}$
        \ENDWHILE
        \end{algorithmic}
        
        \begin{algorithmic}[1]
        \item[]{\bfseries \# Training DNN}
        \FOR{$e = 1, \dots, E$} 
            \FOR{batched pair $\{\boldsymbol{\mathcal{V}},\boldsymbol{\mathcal{Y}}\}$}
                \STATE Train DNN model $\theta^e \gets \text{Adam}(\theta^{e-1}; \{\boldsymbol{\mathcal{V}},\boldsymbol{\mathcal{Y}}\}, \eta, f)$
            \ENDFOR
       \ENDFOR
    \STATE Output the trained DNN model $\theta^\ast \gets \theta^E$
    \end{algorithmic}
\end{algorithm}

\section{Experiment and Results}
\label{Sec. Experiment and Results}
This section is organized as follows to illustrate the experiments and results of this paper. After introducing the non-ideal properties of Velostat and subject information, we conduct two experiments, including human sleeping posture and activity recognition, respectively. A sports game is introduced as an indicator for our activity recognition. The experiment also compares the performances and results of different DNN architectures.

\subsection{Velostat Non-ideal Properties} 
\label{Sec. Velostat Non-ideal Properties}
\begin{figure}[h]
\centering
\subfloat[\label{Fig. Unoccupied Sample Distribution}]{%
  \includegraphics[width=0.5\linewidth]{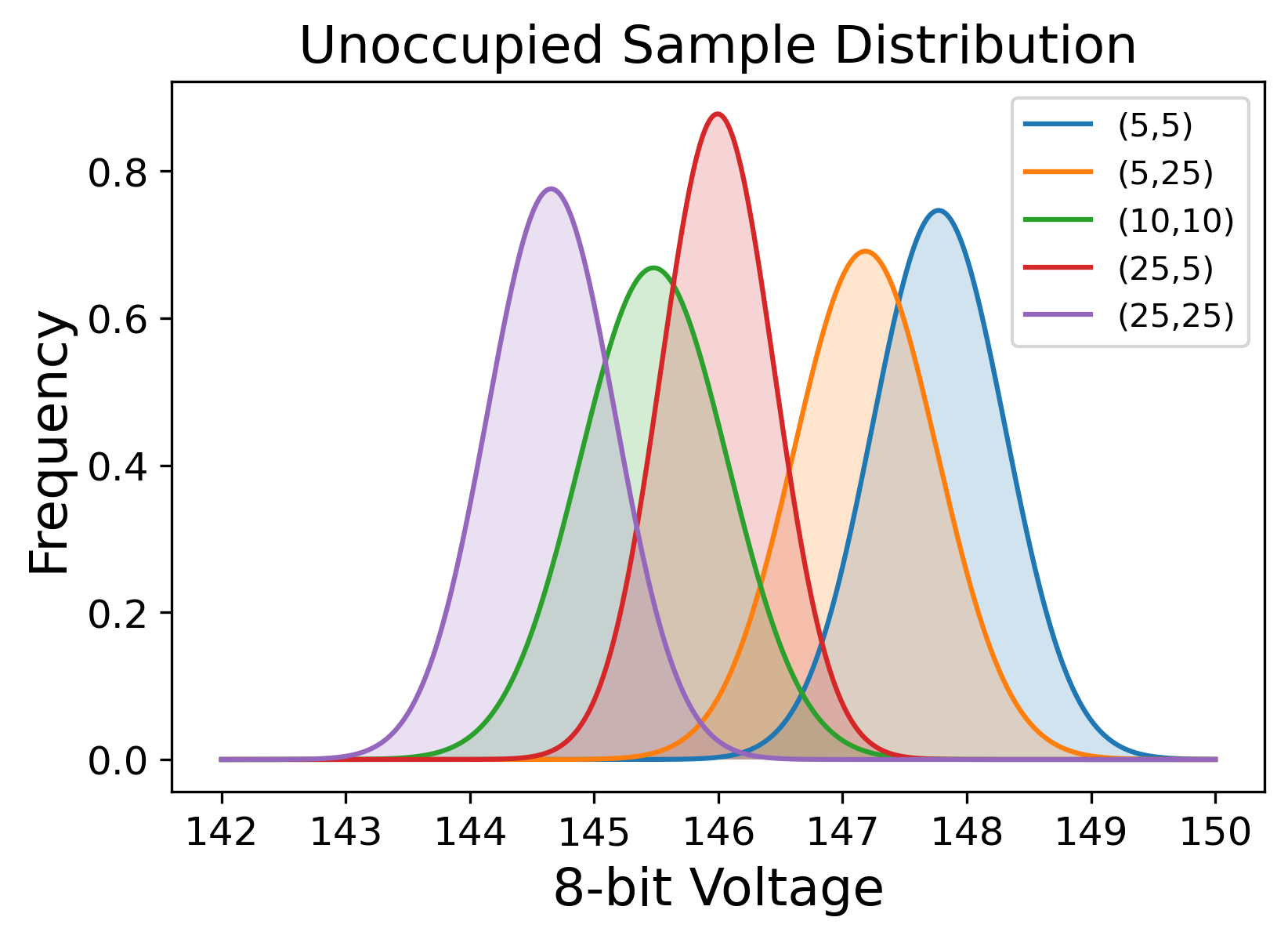}}
\hfill
\subfloat[\label{Fig. Velostat Nonlinear Resistance Characteristics}]{%
  \includegraphics[width=0.5\linewidth]{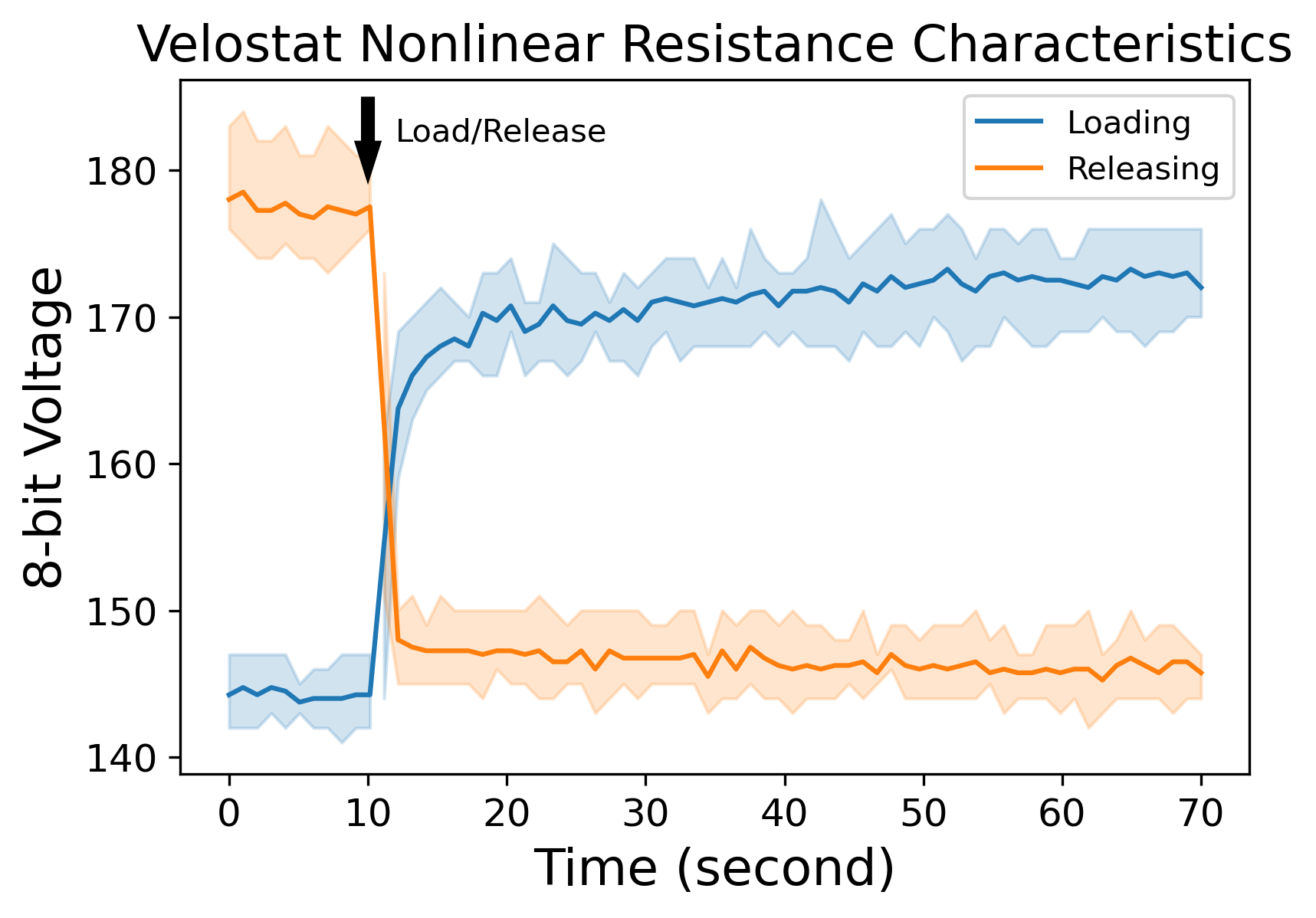}}
\caption{\textbf{Velostat Non-Ideal Properties.} (a) Voltage distribution of five elements when unoccupied. (b) Voltage response curves during loading and releasing states.}
\label{Fig. Velostat Non-ideal Properties} 
\end{figure}

Piezoresistive resistors such as Velostat have some notorious nonideal properties, including reduced sensitivity due to crosstalk, stray currents, electrical noise, and nonlinear resistance characteristics. We seek to improve the accuracy and robustness of SPeM by discussing and characterizing these properties of Velostat and by featuring them in the pressure image stream data. Due to the inevitable crosstalk, the resulting noise is a challenge. Fig. \ref{Fig. Unoccupied Sample Distribution} shows the voltage distribution of different elements when they are not occupied. It can be seen that the voltage of different elements oscillates between several units when unoccupied, which also causes the element voltage to oscillate when monitoring human activities. Fig. \ref{Fig. Velostat Nonlinear Resistance Characteristics} shows the noise-induced voltage instability in the loading and releasing states and the non-linear resistance characteristics. In which the loading and releasing states refer to the state of continuous loading at 100 Newton pressure and the state of after-stop loading, respectively. At the 10-th second, the Velostat is loaded with a voltage rise, but the gradient is decreasing. Although the voltage rise can converge to a final steady-state value, for human monitoring applications, the voltage and the generated pressure image constantly oscillate and rise as the occupancy progresses. The same is true for the releasing state. Therefore, we propose an image stream instead of a single image as a sample in the dataset.

\subsection{Experimental Setup} 
\label{Sec. Experimental Setup}
The data collection is at a home scenario, as it can better fit the functionality of the proposed SPeM product. The portable pressure sensor e-mat that can be easily extended to multiple uses, so it is placed on a mattress and a carpet for different classification tasks. The pressure distribution on e-mat is closely related to the pressure distribution of the human body, and a diverse dataset can effectively avoid overfitting the neural network model. We recruit 14 subjects to assess the feasibility of the SPeM system, comprising 11 males and 3 females. These individuals vary in age, height, and weight. They also have various fitness routines, athletic preferences, and proficiency in certain sports. Detailed physical information about the subjects is presented in Table \ref{Table Subject's physical information}.

\begin{table}[ht] 
\centering
\caption{Subjects' Physical Information, Including Age, Height, Weight, and BMI}
\label{Table Subject's physical information}
\setlength{\tabcolsep}{3pt}
\begin{tabular}{@{}lcccc@{}}
\toprule
 & \textbf{Minimum} & \textbf{Maximum} & \textbf{Average} & \textbf{Standard Deviation}\\
\midrule
Age (years) & 21 & 30 & 23.64 & 3.03 \\
Height (m) & 1.67 & 1.88 & 1.76 & 6.27 \\
Weight (kg) & 62 & 90 & 72.79 & 7.56 \\
BMI & 18.72 & 29.05 & 23.47 & 2.59 \\
\bottomrule
\end{tabular}
\end{table}

Subjects with different instructions completed the posture and activity collection datasets. Ten pressure images are used as a sample of the dataset and have a sampling period of two images per second. The collection time for each sample is approximately five seconds. Appropriate sampling time can not only obtain a higher number of samples but also ensure that each sample contains a sufficient amount of pressure image information to be learned by DNN models. Between samples, subjects moved and rotated consciously to obtain a diverse dataset. The human sleeping postures and dynamic activities are collected in two datasets to validate the application of the proposed SPeM in human monitoring. Each dataset is divided into a training set, a test set, and a validation set with a ratio of 0.7, 0.15, and 0.15, respectively. The optimizer is Adam, with the learning rate initially being $10^{-3}$, and the learning rate decreases by 0.1 every 100 epochs. Since these are classification tasks, the last layer is Softmax, and the loss is estimated by cross-entropy. The batch size is set to 32 and iterated 200 epochs to obtain the preliminary fitting results of the DNN model. The DNN architectures are implemented on TensorFlow and then trained on a Nvidia GeForce RTX(TM) 3080 GPU.

\subsection{Sleeping Posture Recognition} 
\label{Sec. Sleeping Posture Recognition}

\begin{figure}[t]
\centerline{\includegraphics[width=\columnwidth]{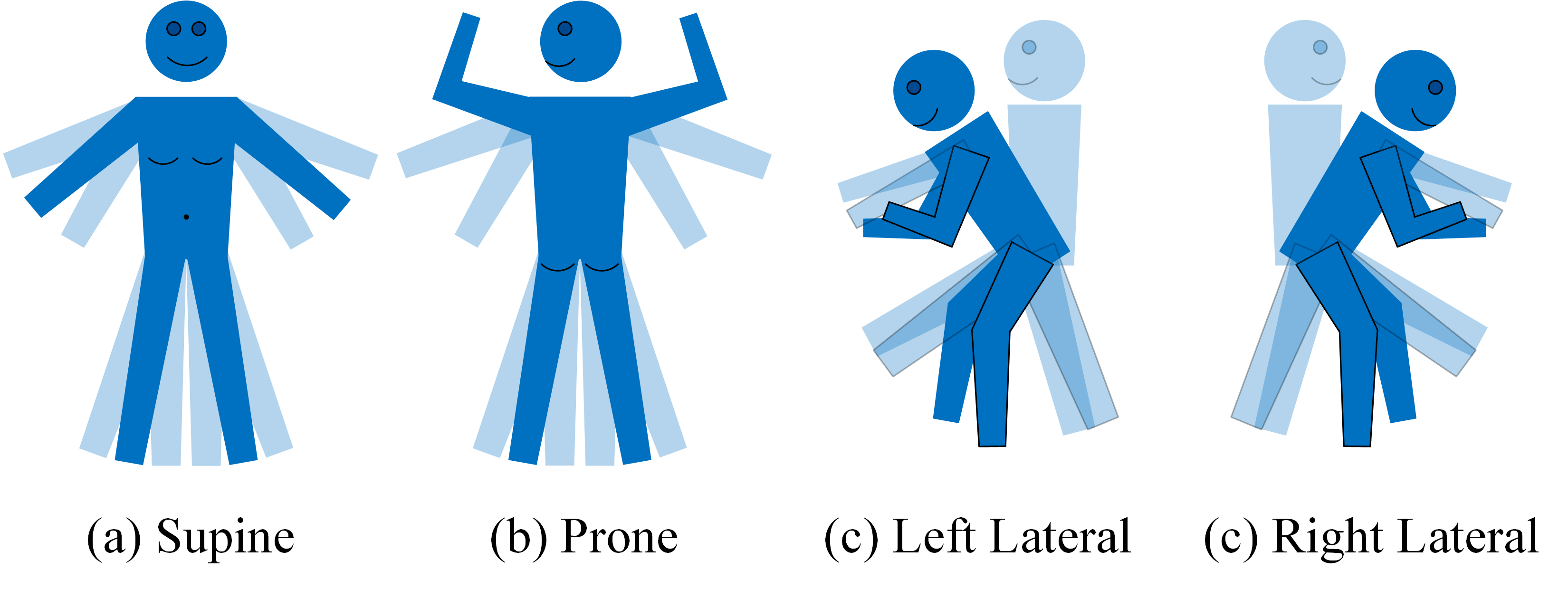}}
\caption{\textbf{Illustration of Four Common Sleeping Postures.} (a) Supine, (b) Prone, (c) Left Lateral, (d) Right Lateral. Bokeh body parts are included in the illustration to represent possible rotation and displacement of the arms, legs, torso, and head during sleep.}
\label{Fig. Human Sleeping Posture}
\end{figure}

SPeM is placed on a queen-size mattress for its first home scenario application, which is the recognition of sleeping postures. This is crucial for high-precision posture recognition that can effectively prevent diseases such as bedsores. As shown in Fig. \ref{Fig. Actual Mat}, SPeM is designed to completely cover the mattress. Subjects are instructed to simulate four common sleeping postures in SPeM, as shown in Fig. \ref{Fig. Human Sleeping Posture}. The subject's body is oriented in the same direction as the mattress, and we allow some degree of movement and rotation of their arms, legs, torso, and head. The orientation of the chest is used to differentiate between the four postures while allowing for some movement and rotation of the subject on the SPeM to diversify the dataset.

\begin{figure}[t]
\centerline{\includegraphics[width=0.8\columnwidth]{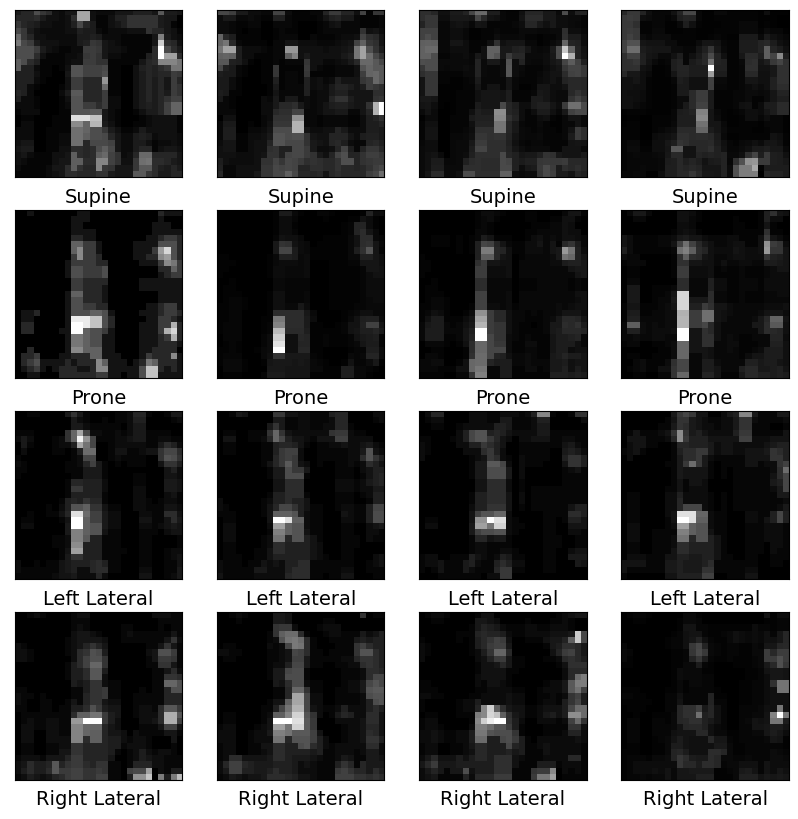}}
\caption{\textbf{Pressure Images of Four Human Sleeping Postures.}}
\label{Fig. Pressure Image Example}
\end{figure}

Fig. \ref{Fig. Pressure Image Example} shows some sampled pressure images of four human sleeping postures, from which a clear outline of the human body and the pressure distribution can be seen. In addition to the changes caused by human pressure, there is also some noise distributed around the edges of the image. Therefore, in the process of recognition of the pressure mode, the classifier should focus more on the center of the pressure image rather than the edge area, and the neural network can undoubtedly achieve this purpose. The classification results of various DNN models are shown in Table \ref{Table. Classification results}, including CRNN, 3DCNN, and CDNN. It can be seen that these three DNN models have all fairly high accuracy and a similar training time. 

\begin{table}[t] 
\centering
\caption{Classification Results of Different DNNs on the Human Sleeping Posture and Dynamic Activity Recognition Datasets}
\label{Table. Classification results}
\begin{tabular}{@{}l|ccc|ccc@{}}
\toprule
\textbf{Task} & \multicolumn{3}{c|}{\textbf{Sleeping Posture}} & \multicolumn{3}{c}{\textbf{Dynamic Activity}} \\
\midrule
Algorithm & CRNN  & 3DCNN & CDNN & CRNN  & 3DCNN & CDNN \\
Accuracy & 0.988 & 0.985 & \textbf{0.993} & \textbf{0.966} & 0.962 & 0.959 \\
Time (s) & 2461 & \textbf{2120} & 2148 & \textbf{5782} & 5829 & 6136 \\
\bottomrule
\end{tabular}
\end{table}

\subsection{Dynamic Activity Recognition} 
\label{Sec. Dynamic Activity Recognition}

To provide a more accurate and vivid description of human activities, we use the successful commercial game product, Nintendo Switch Ring Fit Adventure (RFA) \cite{RFA}, as an indicator in this study. SPeM is considered a significant complementary, prior, and auxiliary sensor to the RFA gaming platform to enhance its performance. We only use the activity definitions, instructions, and scores of the RFA for dynamic activity recognition experiments. The Nintendo Switch console is equipped with various sensors, including an inertial measurement unit (IMU), motion sensing infrared camera, brightness sensors, etc., and can be equipped with a variety of physical games. RFA is an exercise that can help overcome movement disorders and provide therapeutic applications to restore balance and functional mobility, meeting the demand for home exercise \cite{24,wu2022effect,ruth2022acceptance}. By combining two independent Joy-Con controllers equipped with Ring-Con firmware, various fitness movements can be recognized and detected. However, RFA still has a high error rate when differentiating fitness movements, as only the leg side of the sensor set and the Ring-Con sensor set are used to identify all body movements. To improve the efficiency and performance of RFA, we propose using a prior classifier. The pressure system contact sensor can better capture the pressure distribution of the human body and perform high-precision activity recognition.

\begin{figure*}[h]
\centering

\noindent
\begin{minipage}{.05\linewidth}
  \centering
  \rotatebox[origin=c]{90}{Legs}
\end{minipage}%
\begin{minipage}{.95\linewidth}
  \subfloat[Hip Lift]{\includegraphics[width=0.24\linewidth]{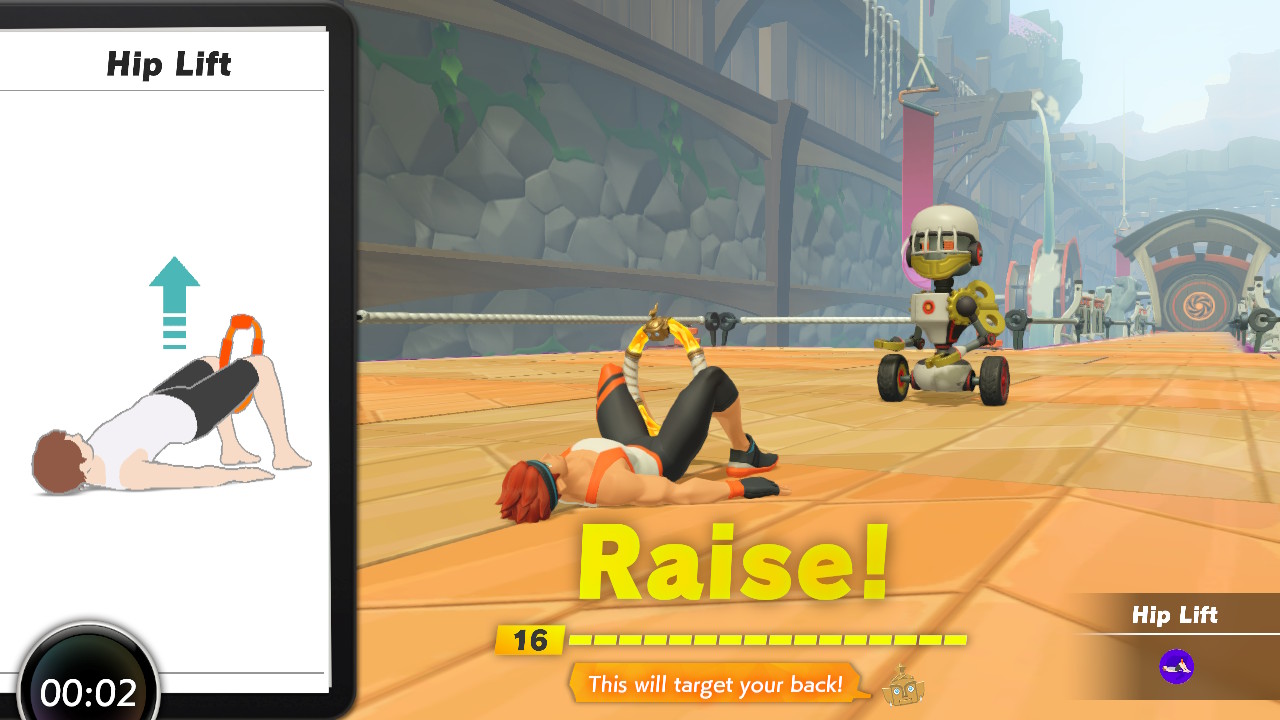}}
  \subfloat[Wide Squat]{\includegraphics[width=0.24\linewidth]{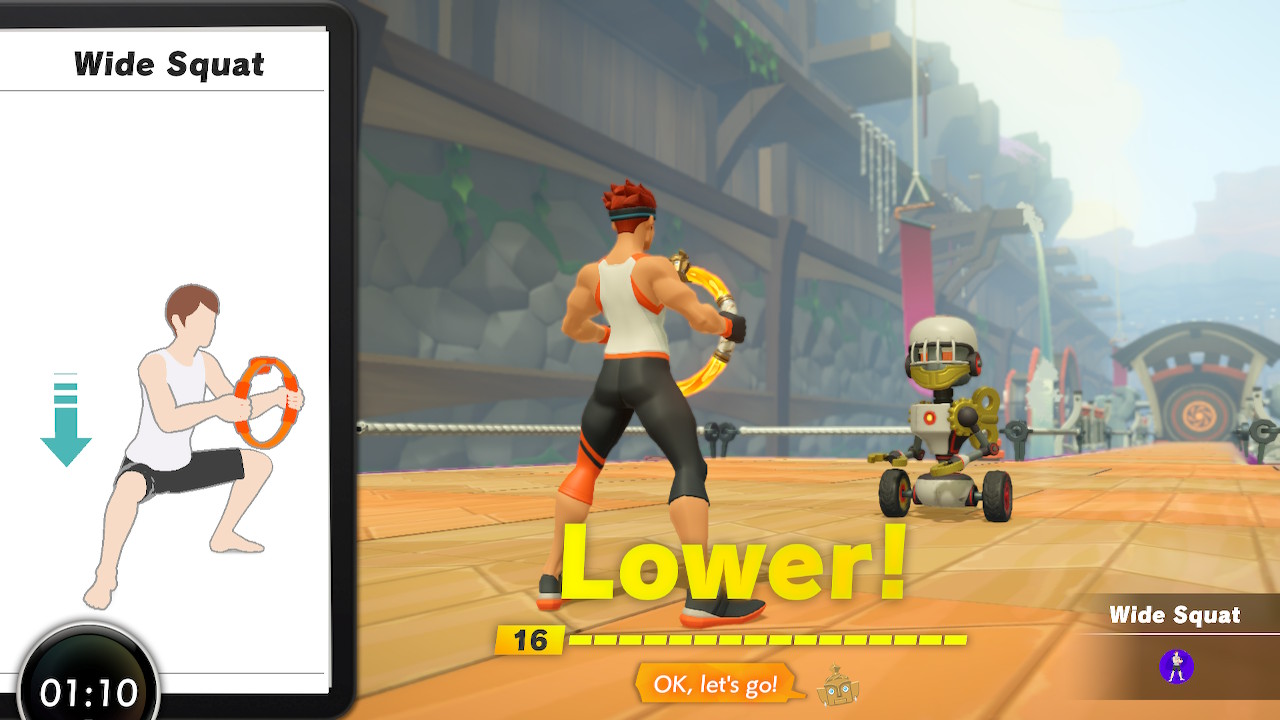}}
  \subfloat[Overhead Squat]{\includegraphics[width=0.24\linewidth]{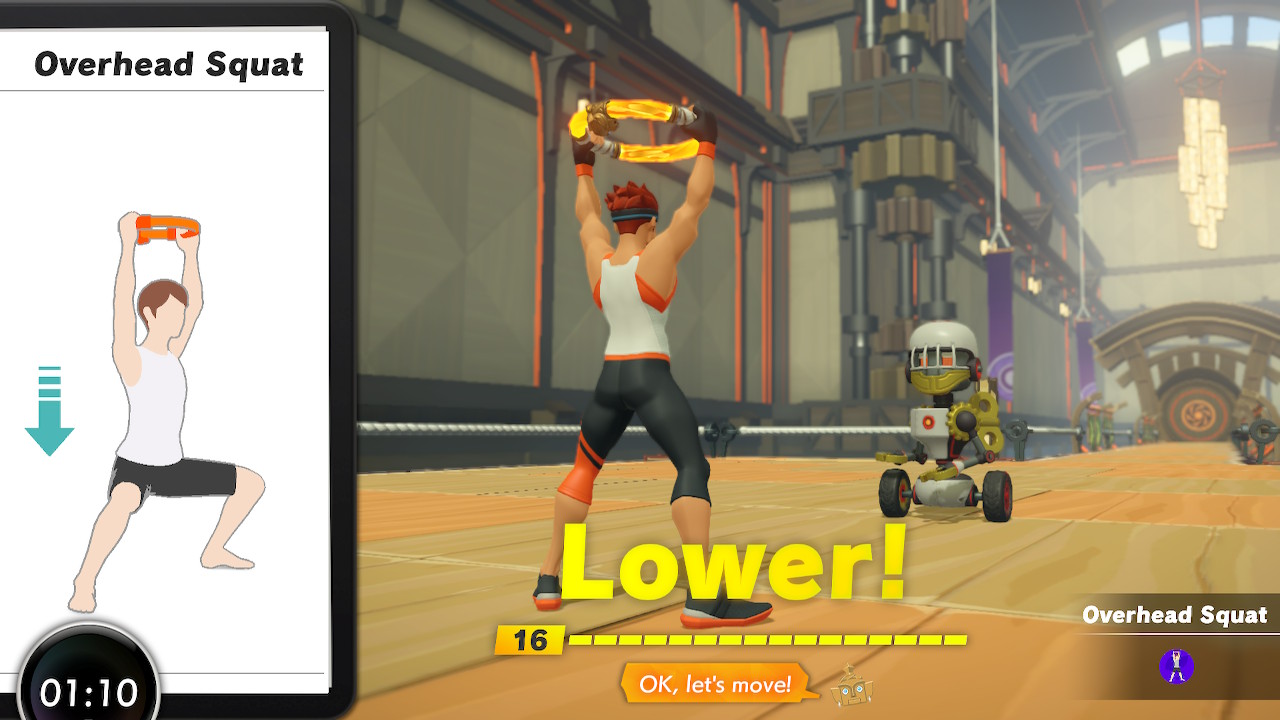}}
  \subfloat[Thigh Press]{\includegraphics[width=0.24\linewidth]{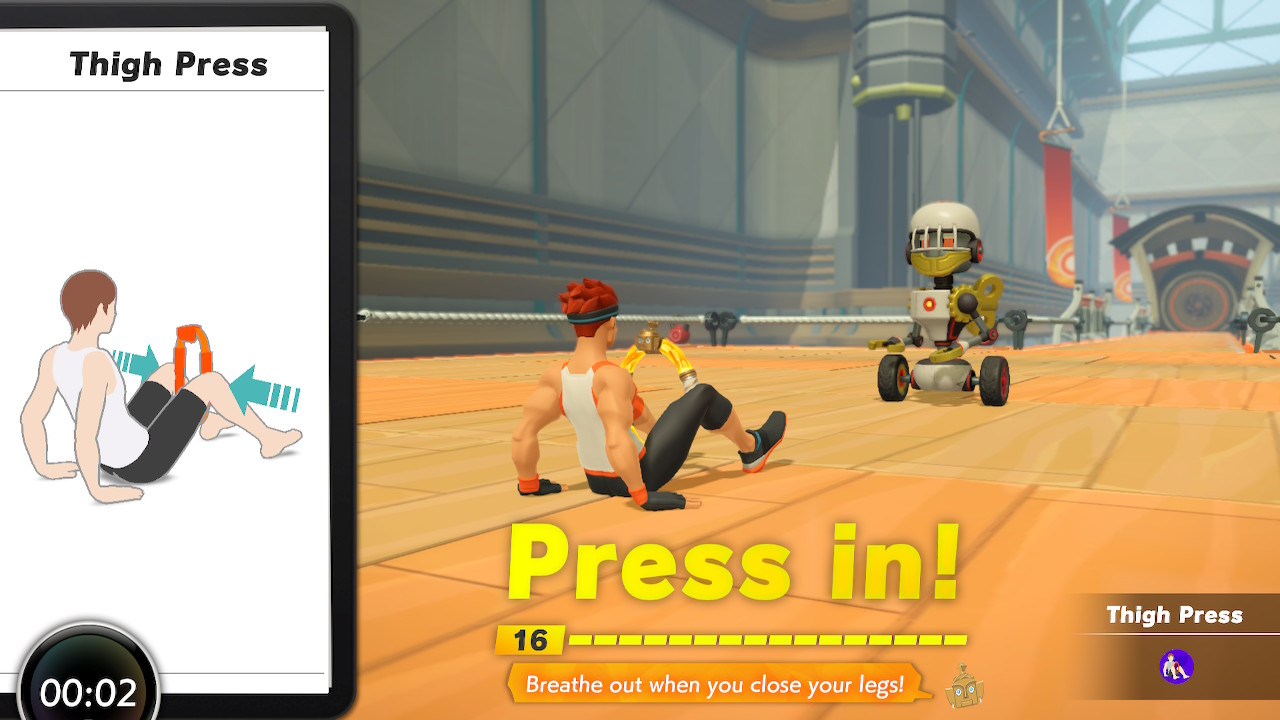}}
\end{minipage}%
\\

\noindent
\begin{minipage}{.05\linewidth}
  \centering
  \rotatebox[origin=c]{90}{Stomach}
\end{minipage}%
\begin{minipage}{.95\linewidth}
  \subfloat[Leg Raise]{\includegraphics[width=0.24\linewidth]{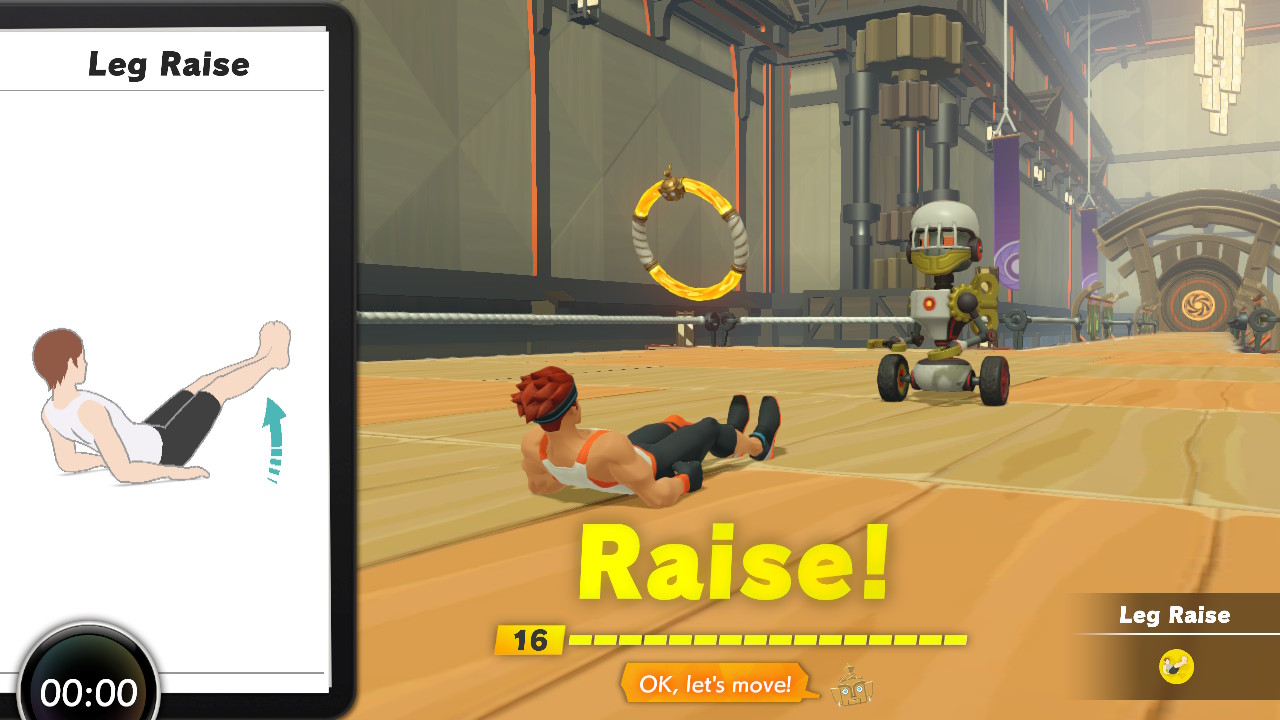}}
  \subfloat[Forward Press]{\includegraphics[width=0.24\linewidth]{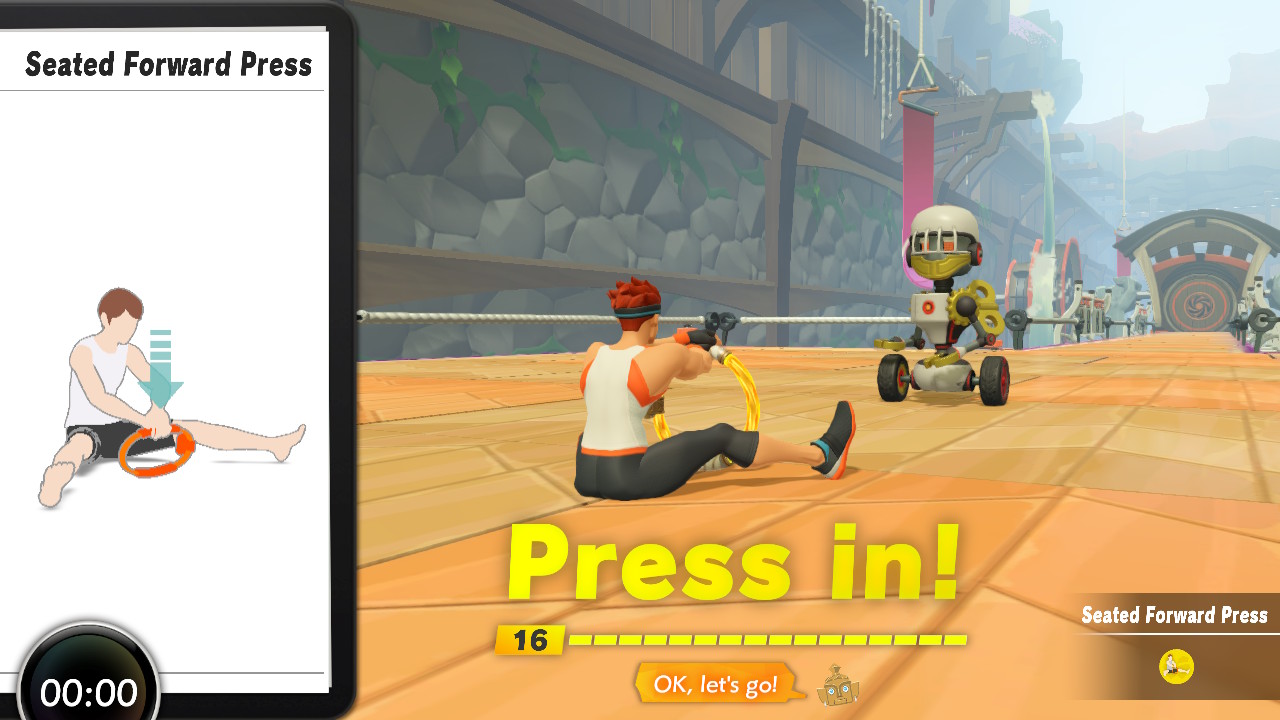}}
  \subfloat[Overhead Bend]{\includegraphics[width=0.24\linewidth]{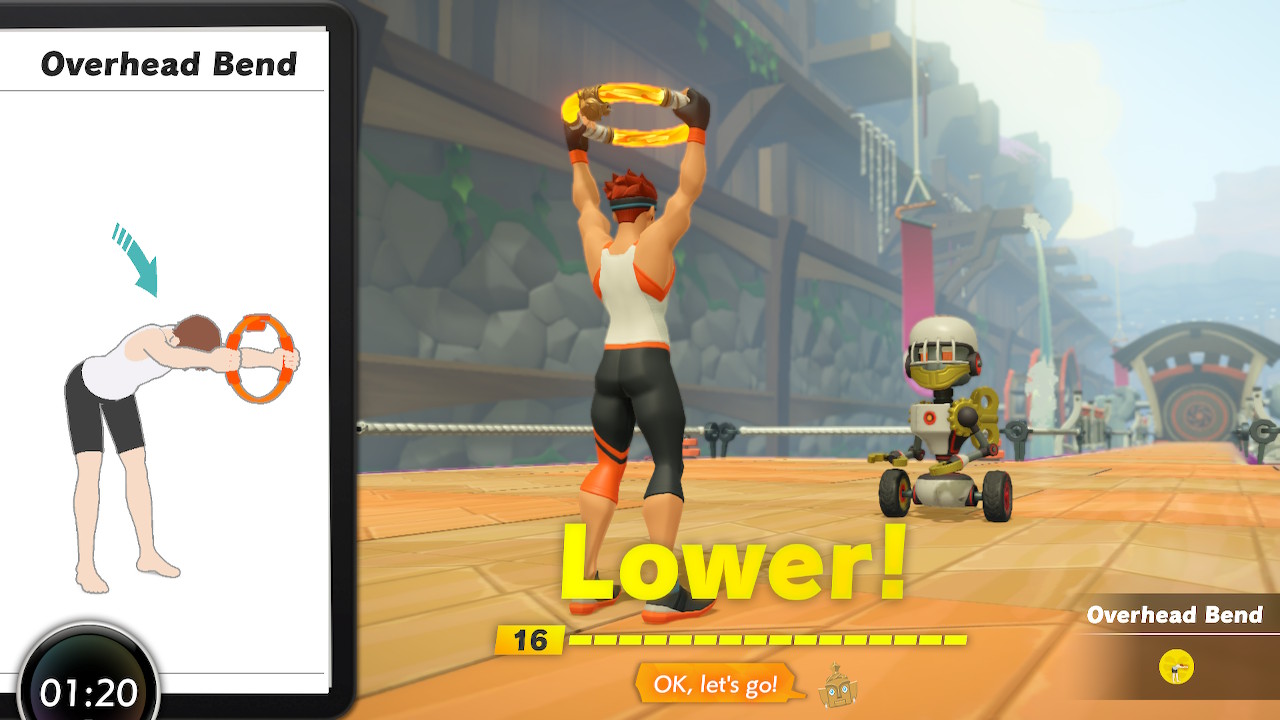}}
  \subfloat[Plank]{\includegraphics[width=0.24\linewidth]{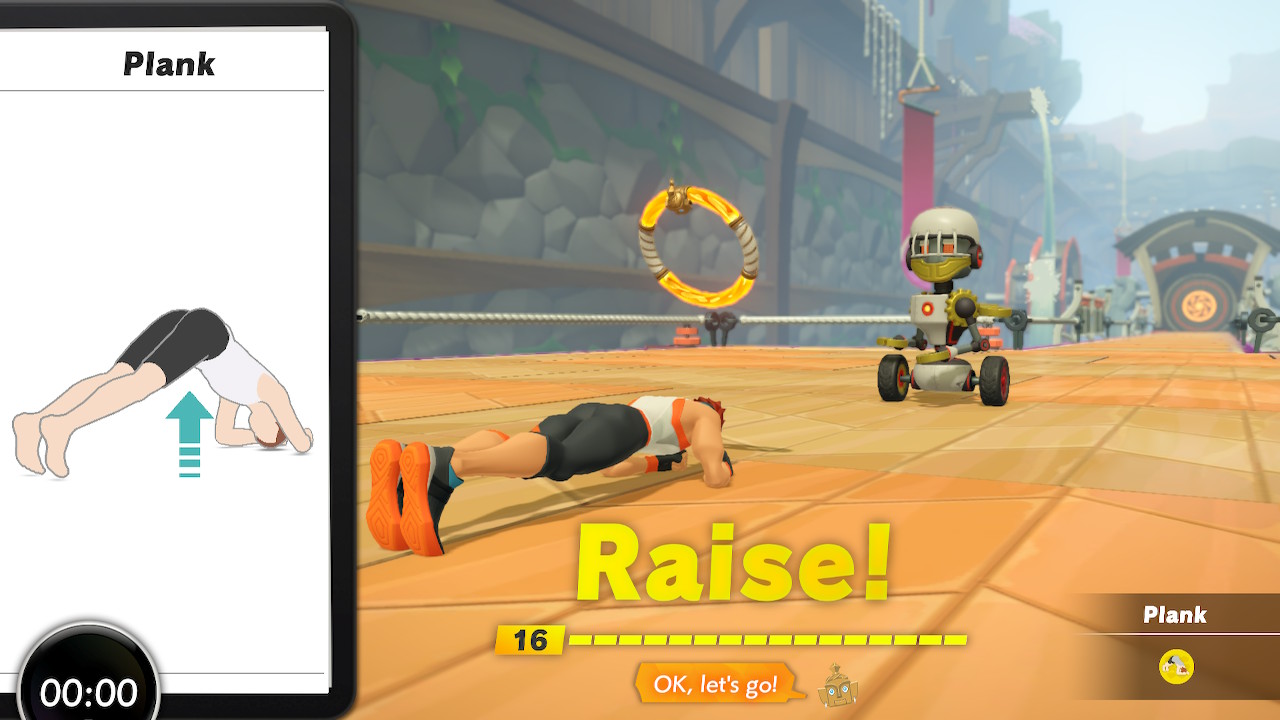}}
\end{minipage}%

\noindent
\begin{minipage}{.05\linewidth}
  \centering
  \rotatebox[origin=c]{90}{Yoga}
\end{minipage}%
\begin{minipage}{.95\linewidth}
  \subfloat[Crescent Lunge]{\includegraphics[width=0.24\linewidth]{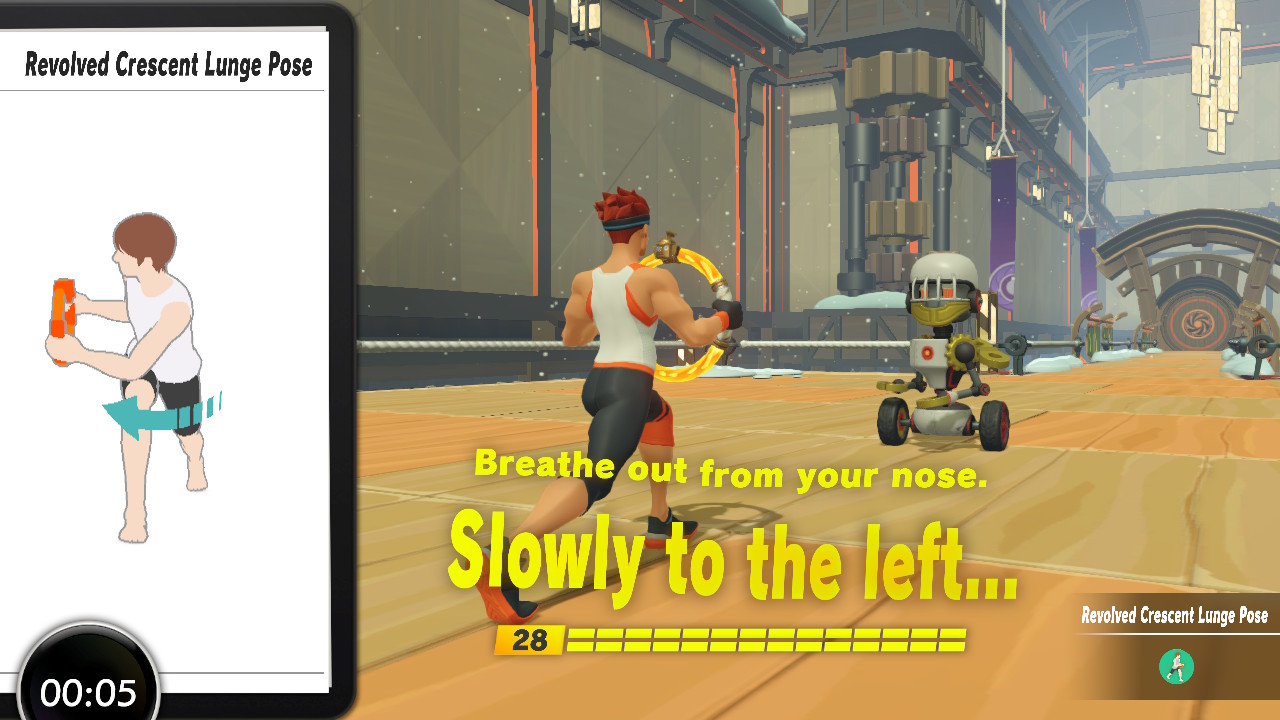}}
  \subfloat[Hinge Pose]{\includegraphics[width=0.24\linewidth]{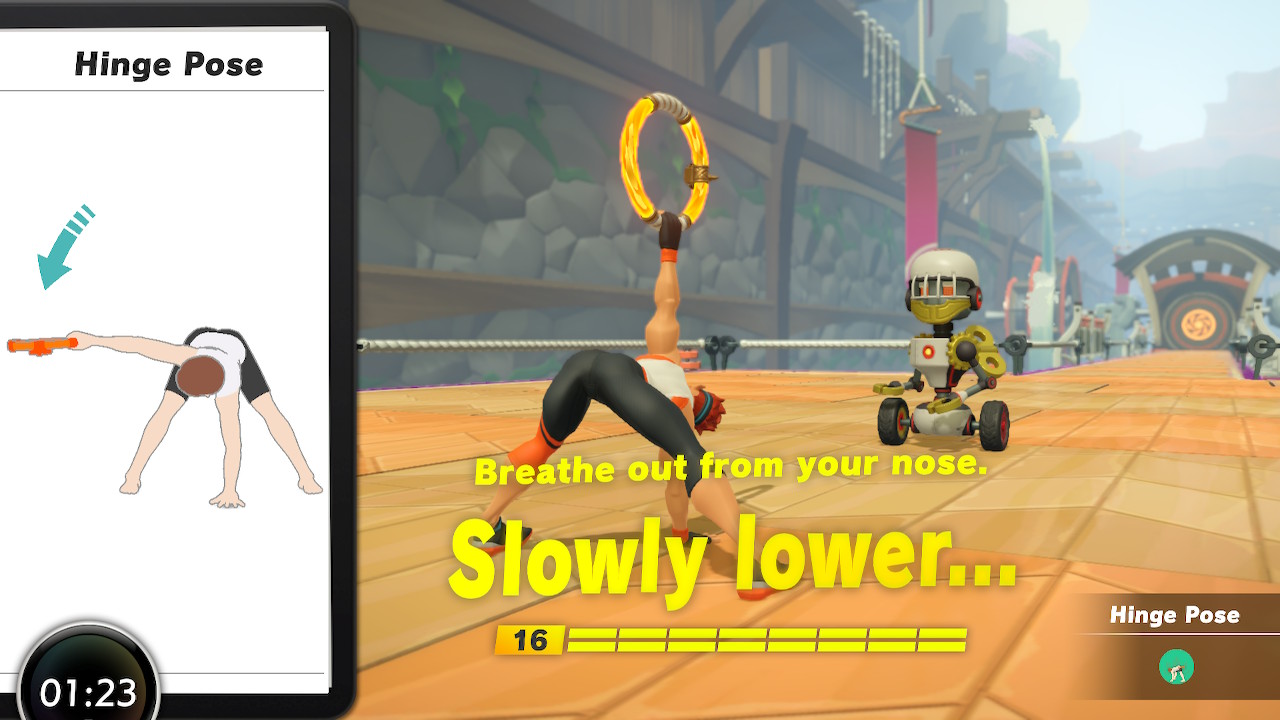}}
  \subfloat[Fan Pose]{\includegraphics[width=0.24\linewidth]{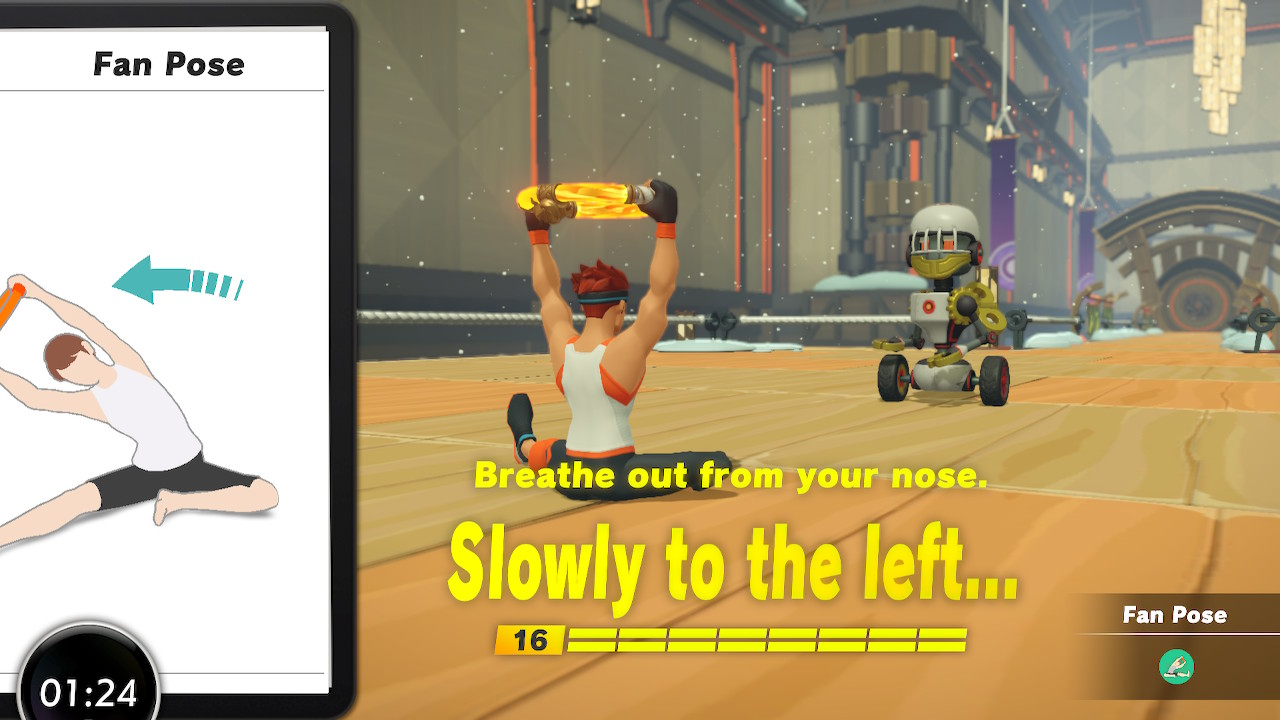}}
  \subfloat[Tree Pose]{\includegraphics[width=0.24\linewidth]{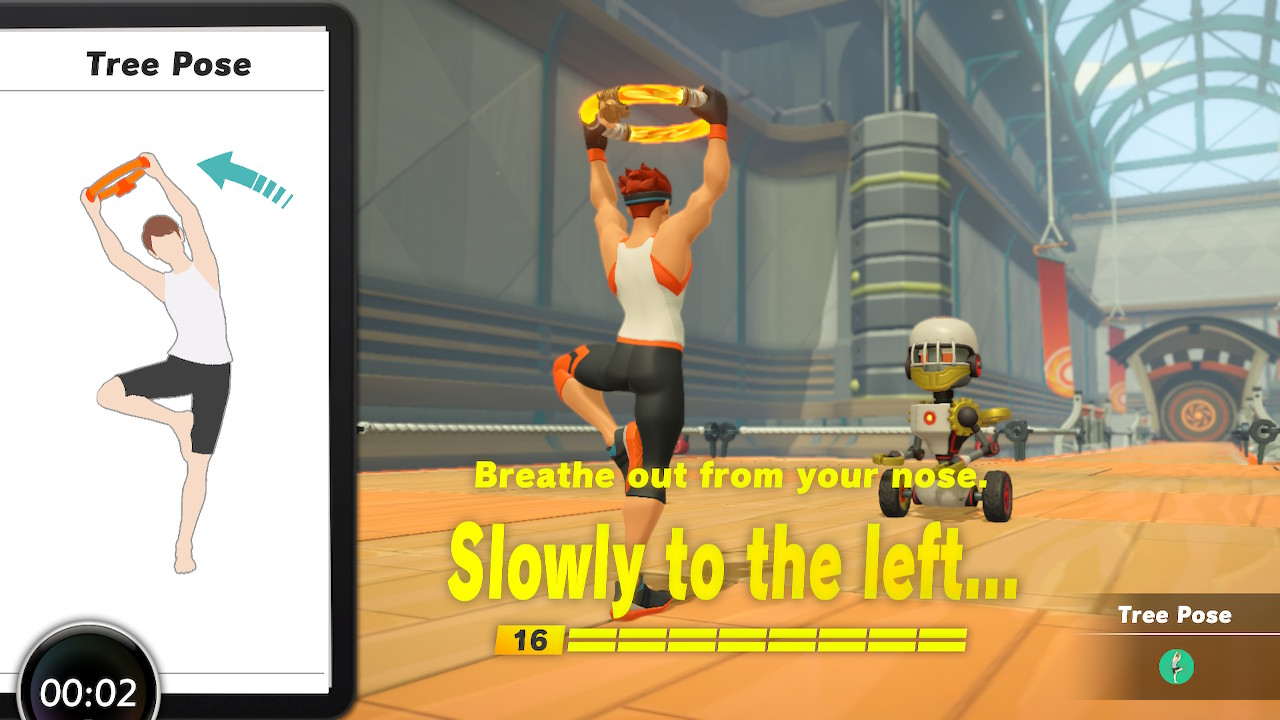}}
\end{minipage}%

\caption{\textbf{Schematic Diagram of Dynamic Activities.} Running is also one of 13 dynamic activities that are not shown here.}
\label{Fig. Dynamic Activity Recognition} 
\end{figure*}

\begin{table*}[t] 
\centering
\caption{Description of the 13 Dynamic Activities}
\label{Table. Description of the 13 Dynamic Activities}
\begin{tabular}{@{}l|l|cccc@{}}
\toprule
\textbf{Category} & \textbf{Activity} & \textbf{Frequency} & \textbf{Amplitude} & \textbf{Size} & \textbf{Contact Point} \\
\midrule
\multirow{4}{*}{Legs} & Hip Lift & Medium & Low & Large & 3 \\
                      & Wide Squat & Medium & Medium & Medium & 2 \\
                      & Overhead Squat & Medium & Medium & Medium & 2 \\
                      & Thigh Press & Low & Medium & Medium & 5 \\
\midrule
\multirow{4}{*}{Stomach} & Leg Raise & Medium & Low & Large & 3 \\
                         & Forward Press & Medium & Low & Large & 2 \\
                         & Overhead Bend & Medium & Medium & Medium & 2 \\
                         & Plank & Low & Medium & Medium & 4 \\
\midrule
\multirow{4}{*}{Yoga} & Crescent Lunge & Low & High & Medium & 2 \\
                      & Hinge Pose & Low & Medium & Medium & 3 \\
                      & Fan Pose & Low & Low & Large & 1 \\
                      & Tree Pose & Low & High & Small & 1 \\
\midrule
Running & Running & High & High & Small & 1 \\
\bottomrule
\end{tabular}
\end{table*}

The design of dynamic tasks is a delicate process, as it requires a link to home sports activities, while also requiring explicit activity instructions and specifications. Therefore, we borrowed activity directives and specifications from the successful case of RFA as the dynamic activity settings for our experiments. The visual cues and scoring system in RFA can effectively reduce the subjects' comprehension errors regarding the experimental instructions, and the RFA can determine whether the subject's activity is normative through a scoring system. RFA includes dozens of activities that work muscles in different parts of the body, such as the chest, waist, and legs. We select 13 dynamic activities to validate the proposed SPeM, as illustrated in Fig. \ref{Fig. Dynamic Activity Recognition}, including (a) Hip Lift, (b) Wide Squat, (c) Overhead Squat, (d) Leg Raise, (e) Forward Press, (f) Overhead Bend, (g) Crescent Lunge, (h) Hinge Pose, (i) Fan Pose, and Running (not shown). RFA categorizes these activities based on training for different body parts, such as Legs, Stomach, and Yoga practices. The pressure patterns generated by these 13 dynamic activities differ in terms of frequency, amplitude, size, and contact points, as summarized in Table \ref{Table. Description of the 13 Dynamic Activities}. For example, the Running activity, characterized by vigorous swinging of both legs, leads to high frequency and high amplitude, and the contact size between the foot and SPeM is relatively small. Note that in addition to these attributes, different activities will produce different shapes, directions, and distances of pressure patterns that originate from various parts of the body. 
We use a non-independent and identically distributed (non-IID) data collection strategy to accurately reflect the real-world preferences of people in their activities. Two subjects collect data on all activities, while five subjects only collect data on certain activities, including (a), (e), (f), (i), and Running. The remaining seven subjects record data for the other eight activities.

\begin{figure}[t]
\centering
  \includegraphics[width=1\linewidth]{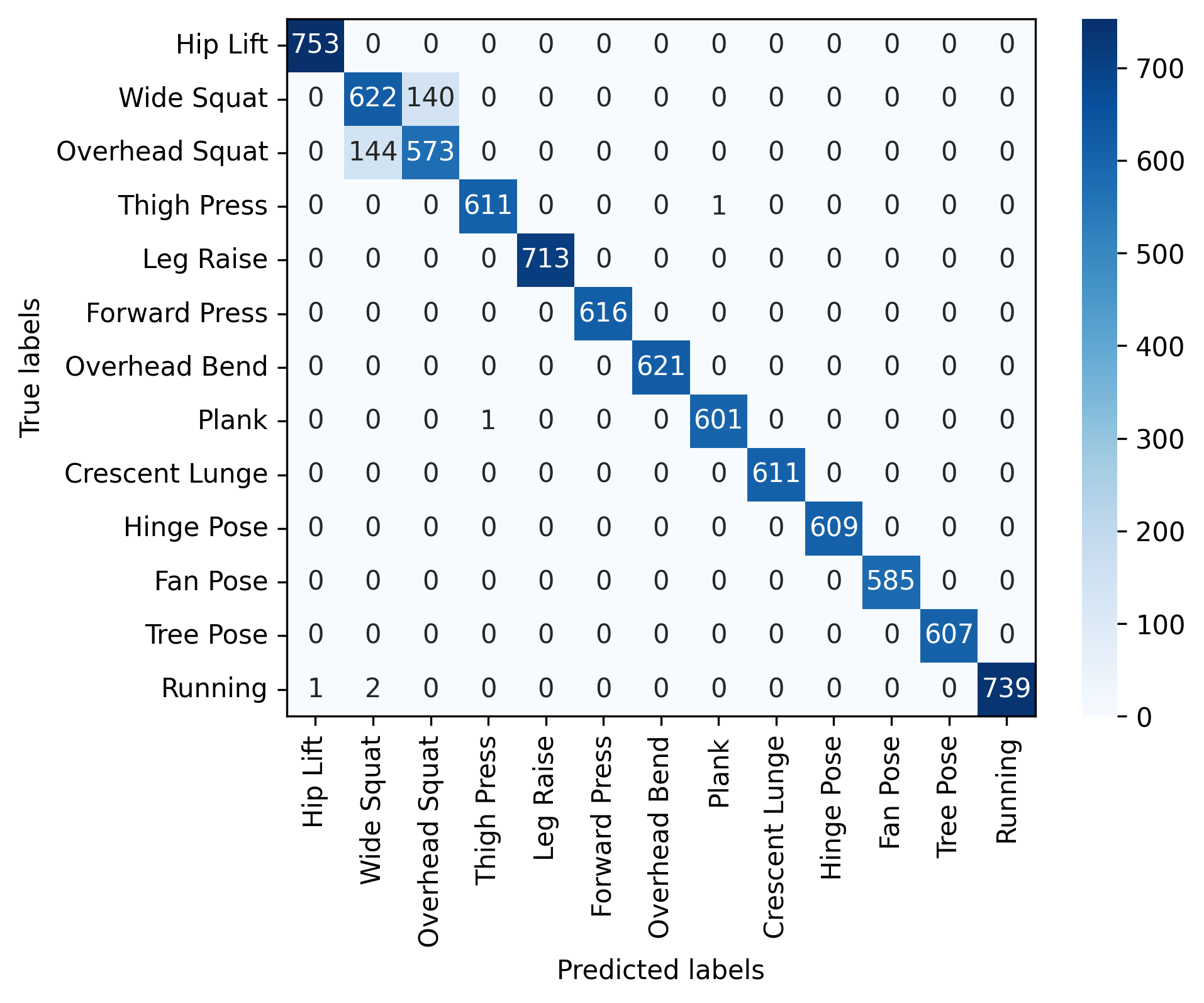}
\caption{\textbf{Confusion Matrix under Predictions of CRNN for 13 Activities Recognition.}}
\label{Fig. Confusion matrix} 
\end{figure}

The results of dynamic activity recognition are shown in Table \ref{Table. Classification results} and Fig. \ref{Fig. Confusion matrix}. All three DNN models achieved similar accuracy in classifying dynamic activities. The primary source of misclassification arises from confusion between the Wide Squat and Overhead Squat activities. These two activities modify the center of gravity by using different arm positions to train leg muscles, but the difference in the pressure modality exerted on the SPeM in the vertical direction is minimal. Therefore, the SPeM can differentiate the pressure modalities of similar activities to some extent, but it is unable to capture or distinguish the behaviors of the head, hands, and torso during activities like Wide Squat and Overhead Squat. To address the observed classification issues and improve the overall accuracy of the system, SPeM could be combined with data from multiple sensors to capture different modalities. For example, combining SPeM with IMU sensors, emblematic of RFAs, might emerge as a promising commercial application avenue. Such multimodal approaches could be essential to address the observed classification challenges and improve the overall accuracy of the system.

\subsection{Effect of Different Image Stream Resolutions}

\begin{figure}[t]
\centerline{\includegraphics[width=\columnwidth]{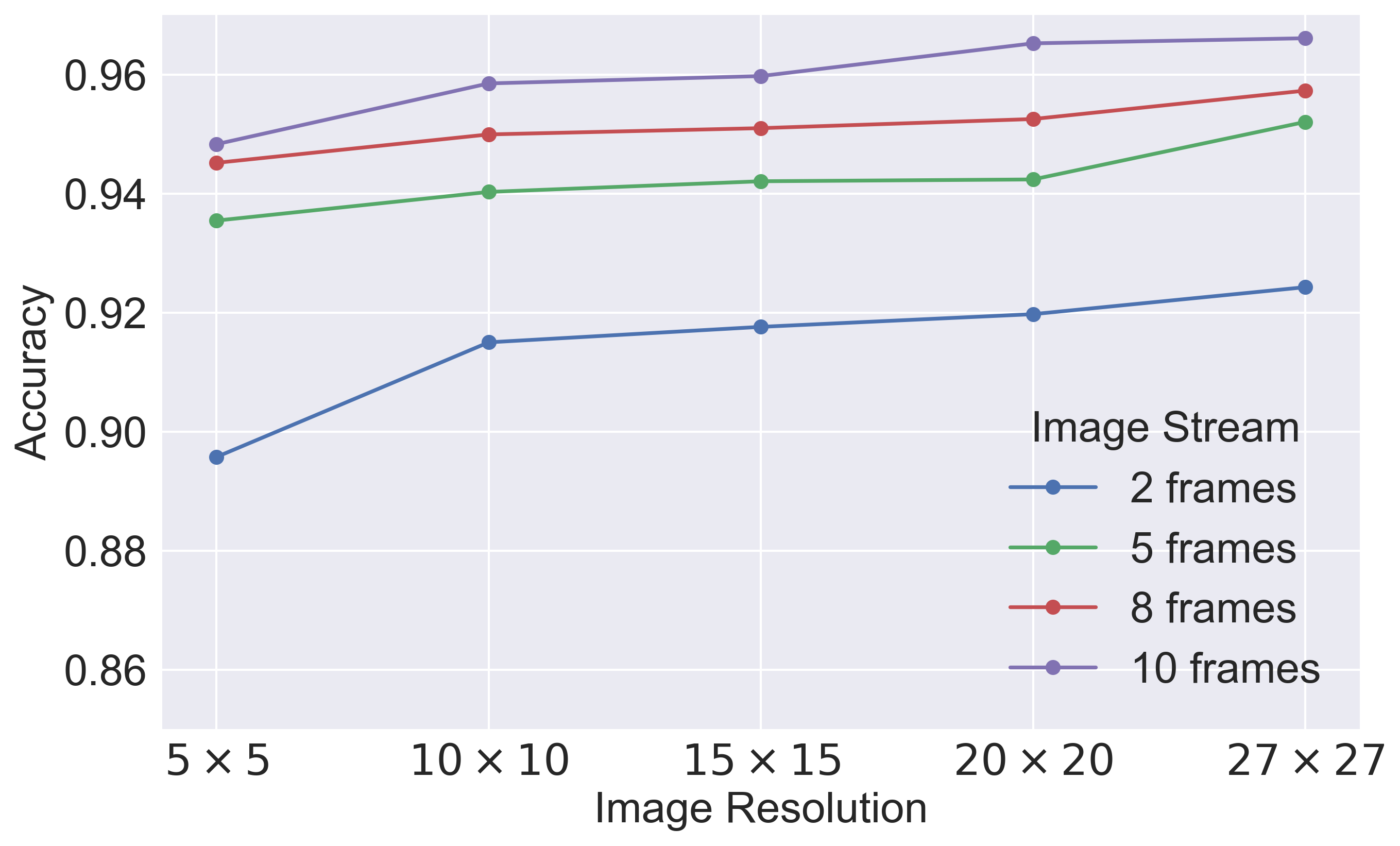}}
\caption{\textbf{Effect of Image Stream Resolution.} The experiment is exemplified with the CRNN for the dynamic activity recognition task.}
\label{Fig. Effect_Resolution}
\end{figure}

To further investigate the impact of different resolutions and image stream frame counts (i.e., sampling rates) on accuracy, we downsample the originally collected samples with a resolution of 27 $\times$ 27 $\times$ 1 $\times$ 10 and retrain and evaluate the DNN, as shown in Fig. \ref{Fig. Effect_Resolution}. We conduct experiments using CRNN for dynamic activity recognition, simulating different resolutions for the SPeM sensor hardware by performing interval sampling. For example, a resolution of 10 $\times$ 10 means sampling every second pixel, resulting in a simulated hardware resolution of 104.6 mm $\times$ 134.6 mm, which is twice the resolution of the original sensor elements shown in Fig. \ref{Fig. Smart Pressure Mat}.

As shown in Fig. \ref{Fig. Effect_Resolution}, while higher resolutions generally lead to better accuracy, there is a diminishing return effect where the rate of improvement slows as the resolution increases. This introduces a trade-off between resolution and system accuracy, which must be carefully considered in different application contexts. Moreover, different use cases may prioritize this trade-off differently. For instance, in healthcare applications such as pressure ulcer prevention, a lower resolution may be sufficient due to the relatively static nature of human postures during sleep, where the cost savings from reduced resolution and sampling rate can outweigh the marginal loss in accuracy. Conversely, in more dynamic applications like home entertainment, where precision and frequent monitoring are essential, a higher resolution and sampling rate may be favored to ensure accurate activity recognition.

\section{Discussion}
\label{Sec. Discussion}

\subsection{Adaptive System Configuration}
The accuracy of the SPeM system will depend on the experimental setup and the resulting data collection process. Two significant factors that affect classification accuracy are the sampling rate and the period. The sampling rate is impacted by the nonlinear resistance characteristics of Velostat. In our experiments, we set the sampling rate to two hertz to account for the application and release of Velostat resistance and the duration of a single activity in the RFA. However, we have observed that the sampling frequency may be a little high, resulting in minor variations in adjacent images and wasting computational resources, as well as electrical noise caused by cluttered currents in the sensor array. It is essential to note that the sampling period depends heavily on the application under consideration. Although the experiments in this paper aim to control the experimental variables and observe the differences between static and highly dynamic applications, a faster sampling period may not be appropriate for other applications, such as sleep recognition, where it could result in the unnecessary waste of computational resources. Therefore, it is necessary to adjust the system parameters to suit different human monitoring applications. A tailored setting of application-following system parameters can lead to better performance and accurate classification results in diverse scenarios.

\subsection{Pressure Image Stream in Real-world Development}
In this paper, we use a DNN algorithm with three architectures to perform the classification task of dynamic pressure image streams. However, based on our observations of related work (Table \ref{Table Comparison}), we have found that most human monitoring applications for static activities, such as sitting, sleeping, and long-term immobility, only use a single image as a data sample. Therefore, we conservatively tested \textsc{ResNet-PI} and other individual CNNs in the sleep posture dataset, achieving a classification accuracy of approximately 0.98. Although single CNNs are a feasible method under ideal conditions, in real-world application scenarios, the proposed image stream-based classification method in the SPeM system is more robust. First, in real-world scenarios, the pressure generated by human sleep is affected by various external factors, such as blankets, pillows, mattresses, and even dolls. Second, the pressure image generated by the Velostat-based pressure sensor array is time-varying due to electrical noise and crosstalk. Third, Velostat resistance also changes over time due to its non-linear resistance characteristics, as shown in Fig. \ref{Fig. Velostat Nonlinear Resistance Characteristics}. For long-term static postures, we can adaptively adjust the sampling rate to reduce system power and storage overheads. Therefore, the introduction of the time dimension can also increase the robustness of human monitoring applications, and we advocate use in static-like human monitoring applications.

\subsection{Future Application Prospects of SPeM}
\label{Sec. Future Application}

The future application prospects of SPeM can be summarized into three aspects: 
\begin{enumerate*}[label=(\roman*)]
    \item serving as a healthcare product under the guidance of professional physicians,
    \item integrating with other sensors to provide more robust performance, and 
    \item further optimizing the design of the sensor array to generate images with higher resolution.
\end{enumerate*}

Firstly, SPeM can be used under the guidance of professional physicians to collect relevant patient data to perform more specialized healthcare tasks, such as preventing bedsores, detecting epilepsy and dementia, identifying falls, among other applications pertaining to the elderly. For example, the high performance of the proposed SPeM in the recognition of posture during sleep suggests its potential to accurately identify the direction of falls in the elderly, providing more precise estimates of the severity of such incidents.

Secondly, SPeM can be fused with other sensors (such as the RFA wearable sensor used in our experiments) to overcome the limitations of a single sensor type. For example, in dynamic activity recognition tasks, the pressure sensor may not fully capture the movement of the human head, torso, or hands. Therefore, combining SPeM with wearable sensors or other sensor types could allow multimodal human monitoring.

Lastly, the proposed SPeM currently generates pressure images with a resolution of $27 \times 27$, which is far from sufficient for visual applications in monitoring human activity. The fabrication of pressure sensor array pads and related applications presents challenges, particularly in large-scale and high-resolution applications. However, because of the law of diminishing returns, increasing size and resolution have limited benefits to recognition accuracy. Beyond employing image-enhancing techniques such as filtering, contrast enhancement, and image interpolation, how to balance cost, material usage, and resolution is an open research question.

\section{Conclusion}
\label{Sec. Conclusion}
In this paper, we propose a SPeM system for large-scale collection and demonstration of human pressure distribution and develop two applications of human sleeping posture and dynamic activity. From the analysis of the pressure image stream, DNN models are used to train, classify, and predict the two datasets. The results of the experiment demonstrate that the proposed SPeM system can excellently achieve high-accuracy classification tasks. This paper highlights the portability, flexibility, and low price of SPeM and demonstrates its preliminary applications in human health care, entertainment, monitoring, etc.

Future work will follow the outline of the prospects for application described in Section \ref{Sec. Future Application}, involving data collection for activity recognition in controlled scenarios, guided by professional physicians and fused with input of multi-sensors. Under the guidance of physicians, subjects could collect data in safer scenarios that mirror professional and realistic situations, such as pressure sores, epilepsy, dementia, falls, etc. Furthermore, we aspire to establish an experimental scenario resembling a hospital ward, deploying pressure sensors in various positions such as rooms, mattresses, and table edges for comprehensive data collection of pressure modes. In this experimental setup, subjects would wear wearable sensors like IMUs for normal living (or simulating diseases), and data labeling would be performed manually using synchronized cameras. The collected dataset will thus form a large-scale pressure modality dataset, providing future researchers with a visual, testable, fusionable, and comparable multimodal dataset with pressure modality as the primary component.

\section*{Acknowledgment}
Thanks to Dr. Erik Blasch and Dr. Robert Ewing for concept development, research discussions, and paper editing.

\section*{References}

\bibliographystyle{IEEEtran}
\small\bibliography{reference}

\begin{thebibliography}{10}
\providecommand{\url}[1]{#1}
\csname url@samestyle\endcsname
\providecommand{\newblock}{\relax}
\providecommand{\bibinfo}[2]{#2}
\providecommand{\BIBentrySTDinterwordspacing}{\spaceskip=0pt\relax}
\providecommand{\BIBentryALTinterwordstretchfactor}{4}
\providecommand{\BIBentryALTinterwordspacing}{\spaceskip=\fontdimen2\font plus
\BIBentryALTinterwordstretchfactor\fontdimen3\font minus \fontdimen4\font\relax}
\providecommand{\BIBforeignlanguage}[2]{{%
\expandafter\ifx\csname l@#1\endcsname\relax
\typeout{** WARNING: IEEEtran.bst: No hyphenation pattern has been}%
\typeout{** loaded for the language `#1'. Using the pattern for}%
\typeout{** the default language instead.}%
\else
\language=\csname l@#1\endcsname
\fi
#2}}
\providecommand{\BIBdecl}{\relax}
\BIBdecl

\bibitem{5}
M.~I. Tiwana, S.~J. Redmond, and N.~H. Lovell, ``A review of tactile sensing technologies with applications in biomedical engineering,'' \emph{Sens. Actuator A Phys.}, vol. 179, pp. 17--31, Jun. 2012.

\bibitem{12}
F.~He, X.~You, W.~Wang, T.~Bai, G.~Xue, and M.~Ye, ``Recent progress in flexible microstructural pressure sensors toward human--machine interaction and healthcare applications,'' \emph{Small Methods}, vol.~5, no.~3, p. 2001041, Mar. 2021.

\bibitem{32}
T.~Al~Shloul, M.~Javeed, M.~Gochoo, S.~Alsuhibany, Y.~Ghadi, A.~Jalal, and J.~Park, ``Student’s health exercise recognition tool for e-learning education,'' \emph{Intell. Autom. Soft Comput}, vol.~35, pp. 149--161, 2022.

\bibitem{2}
M.~Rashid, M.~A. Khan, M.~Alhaisoni, S.-H. Wang, S.~R. Naqvi, A.~Rehman, and T.~Saba, ``A sustainable deep learning framework for object recognition using multi-layers deep features fusion and selection,'' \emph{Sustainability}, vol.~12, no.~12, p. 5037, Jun. 2020.

\bibitem{14}
D.~Aishwarya and R.~Minu, ``Edge computing based surveillance framework for real time activity recognition,'' \emph{ICT Express}, vol.~7, no.~2, pp. 182--186, Jun. 2021.

\bibitem{3}
M.~A. Khan, M.~Mittal, L.~M. Goyal, and S.~Roy, ``A deep survey on supervised learning based human detection and activity classification methods,'' \emph{Multimed. Tools. Appl.}, vol.~80, no.~18, pp. 27\,867--27\,923, Jul. 2021.

\bibitem{yuan2024fedmfs}
L.~Yuan, D.-J. Han, V.~P. Chellapandi, S.~H. Zak, and C.~G. Brinton, ``Fedmfs: Federated multimodal fusion learning with selective modality communication,'' in \emph{ICC 2024-IEEE International Conference on Communications}.\hskip 1em plus 0.5em minus 0.4em\relax IEEE, 2024, pp. 287--292.

\bibitem{yuan2024communication}
L.~Yuan, D.-J. Han, S.~Wang, D.~Upadhyay, and C.~G. Brinton, ``Communication-efficient multimodal federated learning: Joint modality and client selection,'' \emph{arXiv preprint arXiv:2401.16685}, 2024.

\bibitem{attal2015physical}
F.~Attal, S.~Mohammed, M.~Dedabrishvili, F.~Chamroukhi, L.~Oukhellou, and Y.~Amirat, ``Physical human activity recognition using wearable sensors,'' \emph{Sensors}, vol.~15, no.~12, pp. 31\,314--31\,338, Dec. 2015.

\bibitem{nweke2018deep}
H.~F. Nweke, Y.~W. Teh, M.~A. Al-Garadi, and U.~R. Alo, ``Deep learning algorithms for human activity recognition using mobile and wearable sensor networks: State of the art and research challenges,'' \emph{Expert Syst. Appl.}, vol. 105, pp. 233--261, Sep. 2018.

\bibitem{wang2019survey}
Y.~Wang, S.~Cang, and H.~Yu, ``A survey on wearable sensor modality centred human activity recognition in health care,'' \emph{Expert Syst. Appl.}, vol. 137, pp. 167--190, Dec. 2019.

\bibitem{luo2021learning}
Y.~Luo, Y.~Li, P.~Sharma, W.~Shou, K.~Wu, M.~Foshey, B.~Li, T.~Palacios, A.~Torralba, and W.~Matusik, ``Learning human--environment interactions using conformal tactile textiles,'' \emph{Nat. Electron.}, vol.~4, no.~3, pp. 193--201, Mar. 2021.

\bibitem{21}
A.~Voulodimos, N.~Doulamis, A.~Doulamis, and E.~Protopapadakis, ``Deep learning for computer vision: A brief review,'' \emph{Comput. Intell. Neurosci.}, vol. 2018, Feb. 2018.

\bibitem{yuan2023federated}
L.~Yuan, L.~Su, and Z.~Wang, ``Federated transfer-ordered-personalized learning for driver monitoring application,'' \emph{IEEE Internet of Things Journal}, vol.~10, no.~20, pp. 18\,292 -- 18\,301, May 2023.

\bibitem{yuan2023peer}
L.~Yuan, Y.~Ma, L.~Su, and Z.~Wang, ``Peer-to-peer federated continual learning for naturalistic driving action recognition,'' in \emph{Proceedings of the IEEE/CVF Conference on Computer Vision and Pattern Recognition}, 2023, pp. 5249--5258.

\bibitem{zheng2018multispectral}
Y.~Zheng, E.~Blasch, and Z.~Liu, \emph{Multispectral image fusion and colorization}.\hskip 1em plus 0.5em minus 0.4em\relax SPIE Press Bellingham, Washington, 2018, vol. 481.

\bibitem{13}
Z.~Ren, F.~Fang, N.~Yan, and Y.~Wu, ``State of the art in defect detection based on machine vision,'' \emph{Int. J. Precis. Eng. Manuf. - Green Technol.}, pp. 1--31, May 2021.

\bibitem{15}
J.~Chen, J.~Konrad, and P.~Ishwar, ``Vgan-based image representation learning for privacy-preserving facial expression recognition,'' in \emph{Proceedings of the IEEE conference on computer vision and pattern recognition workshops}, 2018, pp. 1570--1579.

\bibitem{yuan2022interpretable}
L.~Yuan, J.~Andrews, H.~Mu, A.~Vakil, R.~Ewing, E.~Blasch, and J.~Li, ``Interpretable passive multi-modal sensor fusion for human identification and activity recognition,'' \emph{Sensors}, vol.~22, no.~15, p. 5787, Aug. 2022.

\bibitem{18}
H.~Li, X.~Zeng, Y.~Li, S.~Zhou, and J.~Wang, ``Convolutional neural networks based indoor wi-fi localization with a novel kind of csi images,'' \emph{China Commun.}, vol.~16, no.~9, pp. 250--260, Sep. 2019.

\bibitem{mu2023human}
H.~Mu, L.~Yuan, and J.~Li, ``Human sensing via passive spectrum monitoring,'' \emph{IEEE Open Journal of Instrumentation and Measurement}, vol.~2, pp. 1--13, September 2023.

\bibitem{4}
M.~Park, B.-G. Bok, J.-H. Ahn, and M.-S. Kim, ``Recent advances in tactile sensing technology,'' \emph{Micromachines}, vol.~9, no.~7, p. 321, Jun. 2018.

\bibitem{chi2018recent}
C.~Chi, X.~Sun, N.~Xue, T.~Li, and C.~Liu, ``Recent progress in technologies for tactile sensors,'' \emph{Sensors}, vol.~18, no.~4, p. 948, Mar. 2018.

\bibitem{23}
\BIBentryALTinterwordspacing
{Adafruit Industries}, ``Pressure-sensitive conductive sheet (velostat/linqstat).'' [Online]. Available: \url{https://www.adafruit.com/product/1361}
\BIBentrySTDinterwordspacing

\bibitem{yuan2021velostat}
L.~Yuan, H.~Qu, and J.~Li, ``Velostat sensor array for object recognition,'' \emph{IEEE Sens. J.}, vol.~22, no.~2, pp. 1692--1704, Dec. 2021.

\bibitem{medrano2019circuit}
C.~Medrano-S{\'a}nchez, R.~Igual-Catal{\'a}n, V.~H. Rodr{\'\i}guez-Ontiveros, and I.~Plaza-Garc{\'\i}a, ``Circuit analysis of matrix-like resistor networks for eliminating crosstalk in pressure sensitive mats,'' \emph{IEEE Sens. J.}, vol.~19, no.~18, pp. 8027--8036, May 2019.

\bibitem{dzedzickis2020polyethylene}
A.~Dzedzickis, E.~Sutinys, V.~Bucinskas, U.~Samukaite-Bubniene, B.~Jakstys, A.~Ramanavicius, and I.~Morkvenaite-Vilkonciene, ``Polyethylene-carbon composite (velostat{\textregistered}) based tactile sensor,'' \emph{Polymers}, vol.~12, no.~12, p. 2905, Dec. 2020.

\bibitem{cao2024crosstalk}
Y.~Cao, Z.~Zhu, M.~Jin, S.~Wang, H.~Shi, P.~Huang, and D.~Hou, ``A crosstalk-free interdigital electrode piezoresistive sensor matrix-based human-machine interaction system for automatic sitting posture recognition,'' \emph{Sensors and Actuators A: Physical}, vol. 371, p. 115284, 2024.

\bibitem{cao2024mortise}
Y.~Cao, Z.~Zhu, S.~Wang, M.~Jin, P.~Huang, and D.~Hou, ``A mortise-tenon structured capacitive pressure sensor array toward large-area indoor activity monitoring,'' \emph{IEEE Sensors Letters}, 2024.

\bibitem{Tactilus}
\BIBentryALTinterwordspacing
``Tactilus® mattress pressure mapping sensor system.'' [Online]. Available: \url{https://tactilus.net/}
\BIBentrySTDinterwordspacing

\bibitem{Tekscan}
\BIBentryALTinterwordspacing
``Body pressure measurement system (bpms).'' [Online]. Available: \url{https://www.tekscan.com/products-solutions/systems/body-pressure-measurement-system-bpms-research}
\BIBentrySTDinterwordspacing

\bibitem{fatema2021low}
A.~Fatema, S.~Poondla, R.~B. Mishra, and A.~M. Hussain, ``A low-cost pressure sensor matrix for activity monitoring in stroke patients using artificial intelligence,'' \emph{IEEE Sens. J.}, vol.~21, no.~7, pp. 9546--9552, Jan. 2021.

\bibitem{sun2017compressed}
C.~Sun, W.~Li, and W.~Chen, ``A compressed sensing based method for reducing the sampling time of a high resolution pressure sensor array system,'' \emph{Sensors}, vol.~17, no.~8, p. 1848, Aug. 2017.

\bibitem{wan2021hip}
Q.~Wan, H.~Zhao, J.~Li, and P.~Xu, ``Hip positioning and sitting posture recognition based on human sitting pressure image,'' \emph{Sensors}, vol.~21, no.~2, p. 426, Jan. 2021.

\bibitem{yuan2021smart}
L.~Yuan and J.~Li, ``Smart cushion based on pressure sensor array for human sitting posture recognition,'' in \emph{2021 IEEE Sensors}.\hskip 1em plus 0.5em minus 0.4em\relax IEEE, 2021, pp. 1--4.

\bibitem{34}
K.~Tang, A.~Kumar, M.~Nadeem, and I.~Maaz, ``Cnn-based smart sleep posture recognition system,'' \emph{IoT}, vol.~2, no.~1, pp. 119--139, Feb. 2021.

\bibitem{26}
Q.~Hu, X.~Tang, and W.~Tang, ``A real-time patient-specific sleeping posture recognition system using pressure sensitive conductive sheet and transfer learning,'' \emph{IEEE Sens. J.}, vol.~21, no.~5, pp. 6869--6879, Dec. 2020.

\bibitem{hudec2020smart}
R.~Hudec, S.~Mat{\'u}{\v{s}}ka, P.~Kamencay, and M.~Benco, ``A smart iot system for detecting the position of a lying person using a novel textile pressure sensor,'' \emph{Sensors}, vol.~21, no.~1, p. 206, Dec. 2020.

\bibitem{sundholm2014smart}
M.~Sundholm, J.~Cheng, B.~Zhou, A.~Sethi, and P.~Lukowicz, ``Smart-mat: Recognizing and counting gym exercises with low-cost resistive pressure sensing matrix,'' in \emph{Proceedings of the 2014 ACM international joint conference on pervasive and ubiquitous computing}, 2014, pp. 373--382.

\bibitem{zhou2017carpet}
B.~Zhou, M.~S. Singh, S.~Doda, M.~Yildirim, J.~Cheng, and P.~Lukowicz, ``The carpet knows: Identifying people in a smart environment from a single step,'' in \emph{2017 IEEE International Conference on Pervasive Computing and Communications Workshops (PerCom Workshops)}.\hskip 1em plus 0.5em minus 0.4em\relax IEEE, 2017, pp. 527--532.

\bibitem{wicaksono20223dknits}
I.~Wicaksono, P.~G. Hwang, S.~Droubi, F.~X. Wu, A.~N. Serio, W.~Yan, and J.~A. Paradiso, ``3dknits: Three-dimensional digital knitting of intelligent textile sensor for activity recognition and biomechanical monitoring,'' in \emph{2022 44th Annual International Conference of the IEEE Engineering in Medicine \& Biology Society (EMBC)}.\hskip 1em plus 0.5em minus 0.4em\relax IEEE, 2022, pp. 2403--2409.

\bibitem{33}
M.~Zhang, D.~Liu, Q.~Wang, B.~Zhao, O.~Bai, and J.~Sun, ``Gait pattern recognition based on plantar pressure signals and acceleration signals,'' \emph{IEEE Trans. Instrum. Meas.}, Sep. 2022.

\bibitem{27}
H.-C. Chen, Y.-K. Jan, B.-Y. Liau, C.-Y. Lin, J.-Y. Tsai, C.-T. Li, C.-W. Lung \emph{et~al.}, ``Using deep learning methods to predict walking intensity from plantar pressure images,'' in \emph{International Conference on Applied Human Factors and Ergonomics}.\hskip 1em plus 0.5em minus 0.4em\relax Springer, 2021, pp. 270--277.

\bibitem{29}
K.~Jun, S.~Lee, D.-W. Lee, and M.~S. Kim, ``Deep learning-based multimodal abnormal gait classification using a {3D} skeleton and plantar foot pressure,'' \emph{IEEE Access}, vol.~9, pp. 161\,576--161\,589, Nov. 2021.

\bibitem{28}
Y.~El~Ghzizal, N.~Aharrane, G.~Khaissidi, and M.~Mrabti, ``Transfer learning and pressure effect for handwriting to early detection of parkinson’s disease,'' in \emph{International Conference on Digital Technologies and Applications}.\hskip 1em plus 0.5em minus 0.4em\relax Springer, 2022, pp. 460--469.

\bibitem{shi2016end}
B.~Shi, X.~Bai, and C.~Yao, ``An end-to-end trainable neural network for image-based sequence recognition and its application to scene text recognition,'' \emph{IEEE Trans. Pattern Anal. Mach. Intell.}, vol.~39, no.~11, pp. 2298--2304, Dec. 2016.

\bibitem{yue2015beyond}
J.~Yue-Hei~Ng, M.~Hausknecht, S.~Vijayanarasimhan, O.~Vinyals, R.~Monga, and G.~Toderici, ``Beyond short snippets: Deep networks for video classification,'' in \emph{Proceedings of the IEEE conference on computer vision and pattern recognition}, 2015, pp. 4694--4702.

\bibitem{song2022contact}
Y.~Song, M.~Li, F.~Wang, and S.~Lv, ``Contact pattern recognition of a flexible tactile sensor based on the cnn-lstm fusion algorithm,'' \emph{Micromachines}, vol.~13, no.~7, p. 1053, Jun. 2022.

\bibitem{7}
S.~Sundaram, P.~Kellnhofer, Y.~Li, J.-Y. Zhu, A.~Torralba, and W.~Matusik, ``Learning the signatures of the human grasp using a scalable tactile glove,'' \emph{Nature}, vol. 569, no. 7758, pp. 698--702, May 2019.

\bibitem{gumus2022textile}
C.~Gumus, K.~Ozlem, F.~Khalilbayli, O.~F. Erzurumluoglu, G.~Ince, O.~Atalay, and A.~T. Atalay, ``Textile-based pressure sensor arrays: A novel scalable manufacturing technique,'' \emph{Micro Nano Eng.}, vol.~15, p. 100140, Jun. 2022.

\bibitem{wai2009sleeping}
A.~A.~P. Wai, K.~Yuan-Wei, F.~S. Fook, M.~Jayachandran, J.~Biswas, and J.-J. Cabibihan, ``Sleeping patterns observation for bedsores and bed-side falls prevention,'' in \emph{2009 Annual International Conference of the IEEE Engineering in Medicine and Biology Society}.\hskip 1em plus 0.5em minus 0.4em\relax IEEE, 2009, pp. 6087--6090.

\bibitem{aoki2014association}
K.~Aoki, M.~Matsuo, M.~Takahashi, J.~Murakami, Y.~Aoki, N.~Aoki, H.~Mizumoto, A.~Namikawa, H.~Hara, M.~Miyagawa \emph{et~al.}, ``Association of sleep-disordered breathing with decreased cognitive function among patients with dementia,'' \emph{J. Sleep Res.}, vol.~23, no.~5, pp. 517--523, Oct. 2014.

\bibitem{matar2019artificial}
G.~Matar, J.-M. Lina, and G.~Kaddoum, ``Artificial neural network for in-bed posture classification using bed-sheet pressure sensors,'' \emph{IEEE J. Biomed. Health Inform.}, vol.~24, no.~1, pp. 101--110, Feb. 2019.

\bibitem{he2016deep}
K.~He, X.~Zhang, S.~Ren, and J.~Sun, ``Deep residual learning for image recognition,'' in \emph{Proceedings of the IEEE conference on computer vision and pattern recognition}, 2016, pp. 770--778.

\bibitem{RFA}
\BIBentryALTinterwordspacing
``Ring fit adventure™ for nintendo switch™ – official site.'' [Online]. Available: \url{https://ringfitadventure.nintendo.com/}
\BIBentrySTDinterwordspacing

\bibitem{24}
T.~B.~F. Pacheco, C.~S.~P. de~Medeiros, V.~H.~B. de~Oliveira, E.~R. Vieira, and F.~De~Cavalcanti, ``Effectiveness of exergames for improving mobility and balance in older adults: A systematic review and meta-analysis,'' \emph{Syst. Rev.}, vol.~9, no.~1, pp. 1--14, Dec. 2020.

\bibitem{wu2022effect}
Y.-S. Wu, W.-Y. Wang, T.-C. Chan, Y.-L. Chiu, H.-C. Lin, Y.-T. Chang, H.-Y. Wu, T.-C. Liu, Y.-C. Chuang, J.~Wu \emph{et~al.}, ``Effect of the nintendo ring fit adventure exergame on running completion time and psychological factors among university students engaging in distance learning during the covid-19 pandemic: Randomized controlled trial,'' \emph{JMIR Serious Games}, vol.~10, no.~1, p. e35040, Mar. 2022.

\bibitem{ruth2022acceptance}
J.~Ruth, S.~Willwacher, and O.~Korn, ``Acceptance of digital sports: A study showing the rising acceptance of digital health activities due to the sars-cov-19 pandemic,'' \emph{Int. J. Environ. Res. Public Health.}, vol.~19, no.~1, p. 596, Jan. 2022.

\end{thebibliography}

\vspace{11pt}

\begin{IEEEbiography}[{\includegraphics[width=1in,height=1.25in,clip,keepaspectratio]{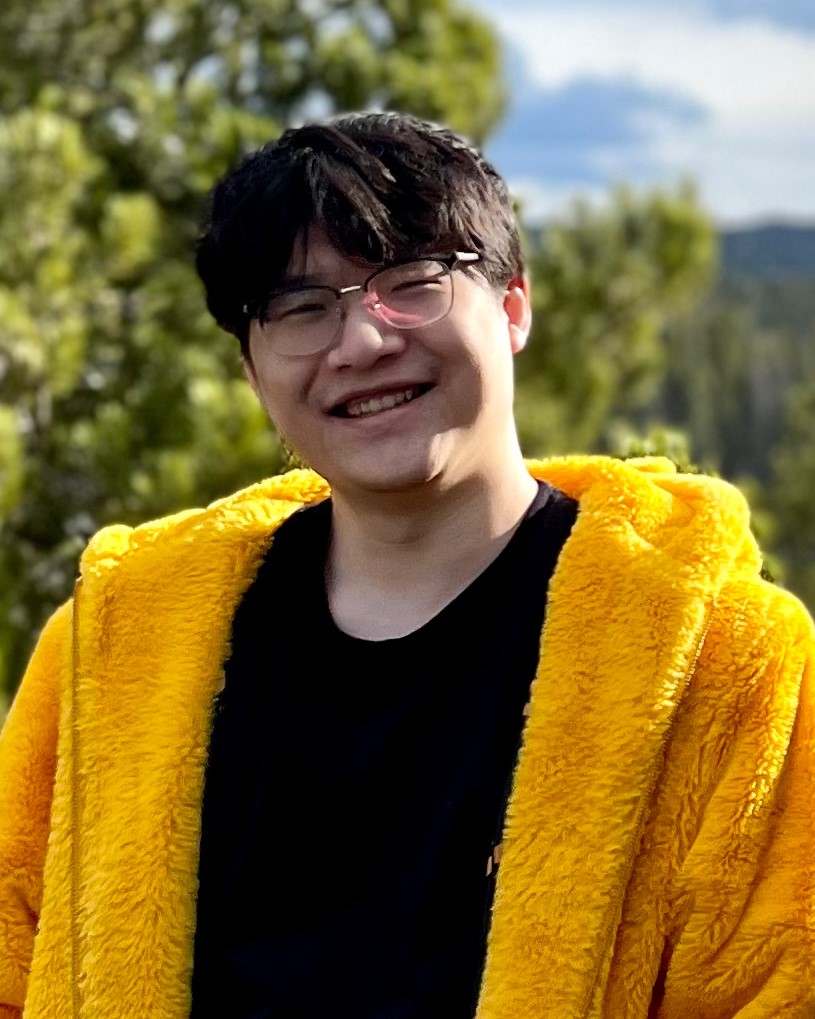}}]
{Liangqi Yuan}
received the B.E. degree from Beijing Information Science and Technology University, Beijing, China, in 2020, and the M.S. degree from the Oakland University, Rochester, MI, USA, in 2022. He is currently pursuing the Ph.D. degree in the School of Electrical and Computer Engineering at Purdue University, West Lafayette, IN, USA. 

His research interests include multimodal learning, mobile computing, and the Internet of Things.
\end{IEEEbiography}

\begin{IEEEbiography}[{\includegraphics[width=1in,height=1.25in,clip,keepaspectratio]{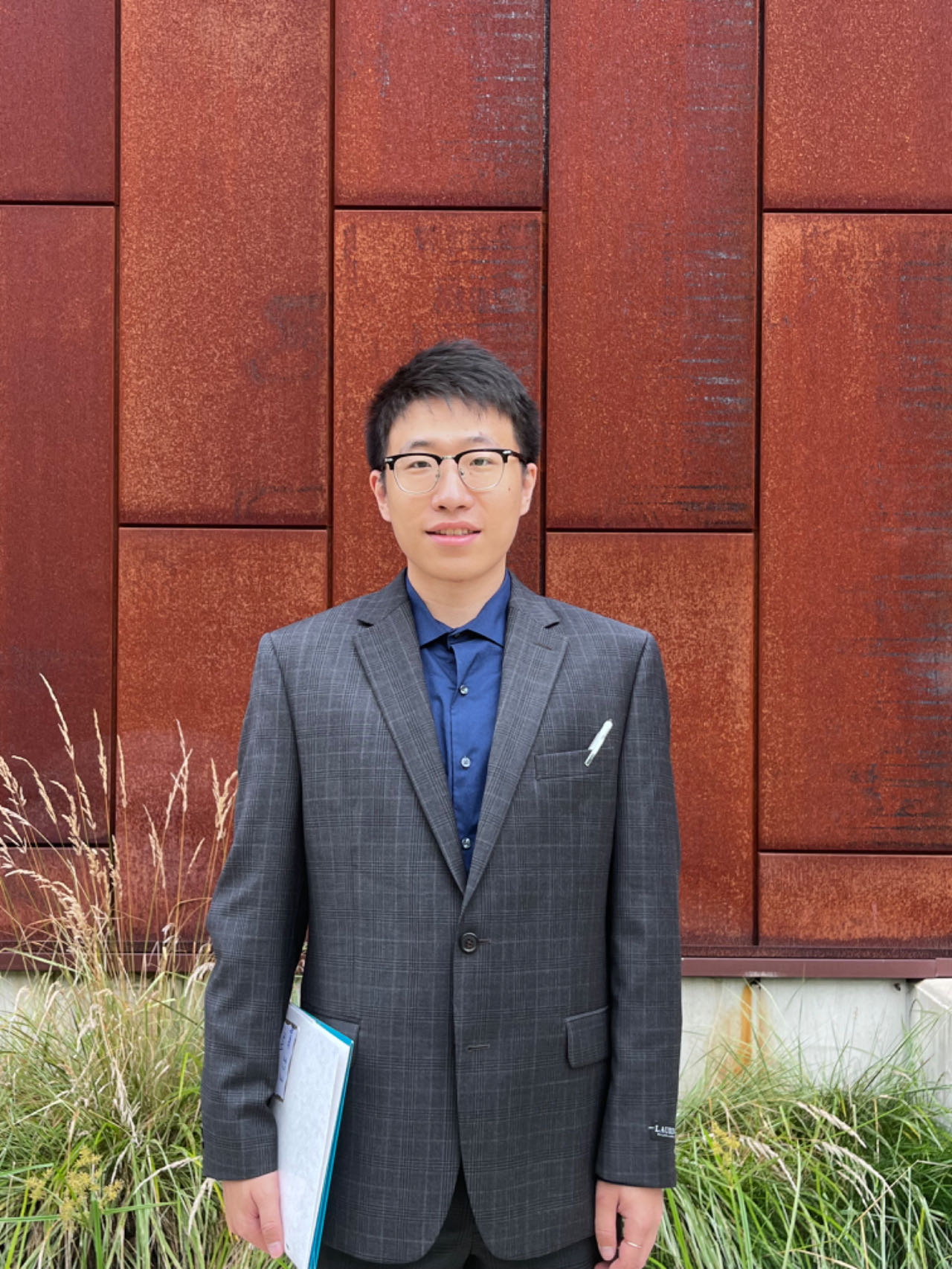}}]
{Yuan Wei}
received B.E degree from Oakland University, Rochester, MI, USA, in 2022, and the B.E. degree from Beijing Information Science and Technoogy University, Beijing, China, in 2023. He is currently pursuing the M.S. degree with the School of Electrical and Computer Engineering, Purdue University, West Lafayette, IN, USA. 

His research interests are in the areas of signal processing, imaging processing, and machine learning. 
\end{IEEEbiography}

\begin{IEEEbiography}[{\includegraphics[width=1in,height=1.25in,clip,keepaspectratio]{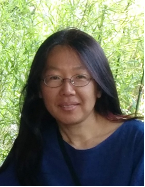}}]
{Jia Li} 
received her B.S. degree in electronics and information systems from Peking University, Beijing, China, in 1996, the M.S.E. degree, and the Ph.D. degree both in electrical engineering from the University of Michigan, Ann Arbor, MI, in 1997 and 2002, respectively. She has been a faculty member in the School of Engineering and Computer Science at Oakland University since 2002. 

Her research interests are in the areas of statistical learning and signal processing with applications in radar, sensor fusion, communications and biomedical imaging. Dr. Li has authored/co-authored over one hundred referred publications, including one book. Her past and current researches are sponsored by NSF, NIH, General Motors, Fiat Chrysler, Ford, National Research Council and Air Force Office of Scientific Research. Dr. Li serves as a member of technical committees of several international conferences and workshops, and a regular reviewer of a number of international journals. She is a senior member of IEEE.
\end{IEEEbiography}

\vfill

\end{document}